\documentclass{article}

\usepackage{PRIMEarxiv}

\usepackage[utf8]{inputenc} 
\usepackage[T1]{fontenc}    
\usepackage{hyperref}       
\usepackage{url}            
\usepackage{booktabs}       
\usepackage{amsfonts}       
\usepackage{nicefrac}       
\usepackage{microtype}      
\usepackage{lipsum}
\usepackage{fancyhdr}       
\usepackage{graphicx}       
\graphicspath{{media/}}     

\usepackage{amsmath}
\usepackage{bbold}
\usepackage{algorithm}
\usepackage{algpseudocode}
\usepackage{subcaption}
\usepackage{multirow}

\title{Reverse Back Propagation to Make Full Use of Derivative
}

\author{
  WeimingXiong \\
  Nanjing University \\
  \texttt{weimingxiong@smail.nju.edu.cn} \\
   \And
  RuoyuYang \\
  Nanjing University \\
  \texttt{yangry@nju.edu.cn} \\
}

\begin{document}
\maketitle

\begin{abstract}
The development of the back-propagation algorithm represents a landmark in neural networks. We provide an approach that conducts the back-propagation again to reverse the traditional back-propagation process to optimize the input loss at the input end of a neural network for better effects without extra costs during the inference time. Then we further analyzed its principles and advantages and disadvantages, reformulated the weight initialization strategy for our method. And experiments on MNIST, CIFAR10, and CIFAR100 convinced our approaches could adapt to a larger range of learning rate and learn better than vanilla back-propagation.

\end{abstract}


\section{Introduction}

Supervised learning in multi-layered neural network through the well-known back-propagation algorithm is the current mainstream neural network training paradigm in deep learning. The error back-propagation learning consists of two passes: a forward pass and a backward pass. In the first pass, the (input) signal is forwarded layer-wisely. Finally, a set of outputs is produced as a response of the neural network. In the second pass, an error signal originating at the output end is propagated backward through the network and finally reaches the input end.	

\begin{figure}[h]
	\centering
	\includegraphics[width=0.8\textwidth]{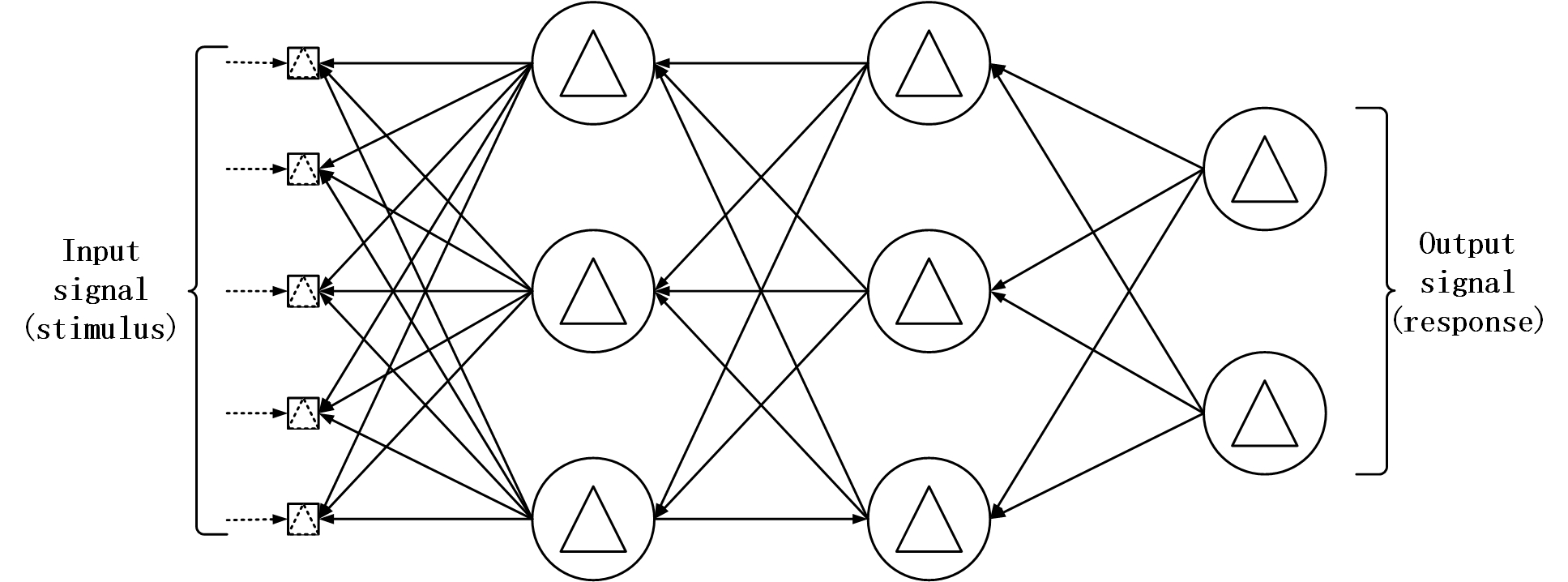}
	\caption{The error back-propagation process of a multi-layer perceptron. We use squares to represent input neurons to distinguish them from neurons. The triangle in the node represents the error related to the derivative of the error function. For clarity, we omitted the errors of synaptic weight. The error signal propagates layer by layer from back to front, and eventually ends at the input end. Graph modified from \cite{haykin1999neural}.}
	\label{demo}
\end{figure}

The multi-layer perceptron in figure \ref{demo} consists of three parts, input neurons, neurons and synaptic weights, each of which contributes differently to the output. During the first pass, the input neuron is responsible for sensing the external input signal, and will not process the signal, while the neuron combines those signals from other neurons connected to its synapses before applying non-linearity. In addition, these three parts behave differently in the second pass. The synaptic weight will retain the error and update itself. The neuron will not retain the error but propagate the error backward. The input neuron is only responsible for sensing the input signal from the outside and does not process the signal in the first pass. Therefore, without participating in the second pass, the back-propagation learning stops here.

In the second pass of back-propagation learning, the error of the input neuron (dashed triangle in figure \ref{demo}) is not calculated. This is reasonable in supervised learning. Because the error surface is only related to the synaptic weights, all input-output cases are fixed and may even be finite, and we cannot optimize the target function by varying something of the input. But we think this is a waste of derivative. Moreover, the error of the input neuron would accumulate as the error back-propagation learning process progresses, and we did some experiments to prove our assumption. This is a paradox, but taking the error of the input neuron into consideration is necessary.

\begin{figure}[h]
	\centering
	\includegraphics[width=0.8\textwidth]{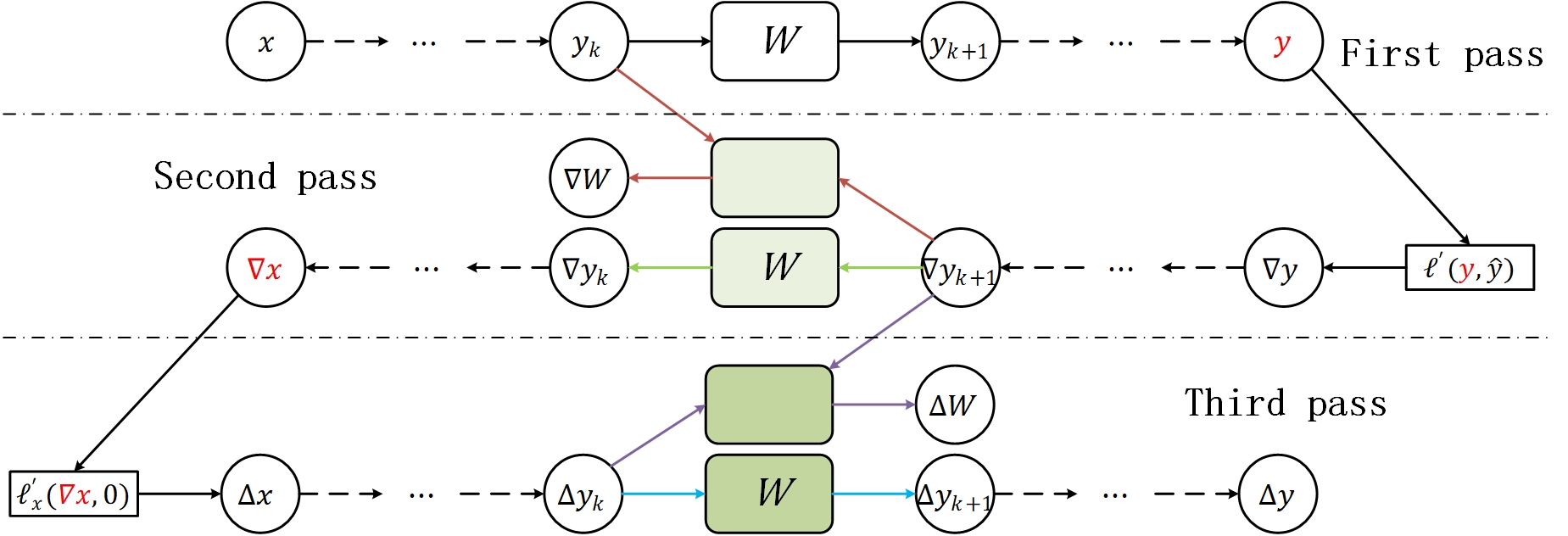}
	\caption{Three passes of a multi-layered neural network. $x, y$ and $\hat{y}$ are the input, output and the desired output of the network. The rounded rectangle represents the operation or transformation of that layer with parameter $W$(empty means parameter-free).$\nabla *$ and $\Delta *$ are gradients (partial derivatives) with respect to $\ell$ and $\ell_x$ respectively. Each layer in the three passes shares the parameter $W$.}
	\label{abp overview}
\end{figure}

In this paper, we introduce the third pass to make full use of derivative wasted in the second pass. Inspired by \cite{glorot2010understanding, KaimingHe2015}, we expect the sum of the variance of $\Delta y_i$ and $\nabla y_i$ in figure \ref{abp overview} in each layer to be the same to avoid gradient diminishing or explosion, and we came to almost the same conclusion as the original paper finally, which further proves the rationality of our method. Our experiments show that the three passes learning surpasses the vanilla back-propagation learning performance on MNIST, CIFAR-10 and CIFAR-100.
At last, we emphasize that $\Delta *$ in figure \ref{abp overview} is not a second derivative, and there is no guarantee that our method can get rid of the local minima.




\section{Background}
\label{sec:background}

The back-propagation algorithm used in deep learning and the advances in hardware, especially Nvidia GPUs that support high-performance parallel computing, have contributed to the overwhelming success of deep learning techniques. It’s hard for us to imagine a neural network with millions of parameters as a function, but a computer is competent for that. 

Back-propagation has a colorful history. \cite{118638} presents a survey of the basic theory of the back-propagation neural network which shows that back-propagation was originally introduced in \cite{bryson1969applied} and independently rediscovered in \cite{Werbos1974}, \cite{parker1985learning, parker1986comparison} and \cite{Rumelhart1986, anderson1988neurocomputing}. \cite{Rumelhart1986} describes the back-propagation learning procedure for networks of neuron-like units, proposes its use for machine learning, and demonstrates how it could work. And the published book \cite{mcclelland1986parallel} has been a major influence in the use of back-propagation learning, which emerged as the most popular learning algorithm for the training of multi-layer perceptrons. The vanishing gradient and other problems exist from then on. To this end, \cite{hinton2006reducing} proposed a way could make neural networks that were well initialized and fine-tuned through back-propagation deeper. Since then, deep learning equipped with back-propagation gradually recovered again.

There are some works considering the local minima caused by back-propagation\cite{107014, baldi1989neural,Sussmann1997,Weixing2005}. In \cite{107014}, some conditions on the network architecture and the learning environment, which ensure the convergence of the BP algorithm, are proposed. Recent work \cite{metz2021gradients} discuss a common chaos based failure mode which appears in a variety of differentiable circumstances, ranging from recurrent neural networks and numerical physics simulation to training learned optimizers, and conclude that gradients are not all you need. \cite{xu2021raise} proposes a straightforward yet effective fine-tuning technique, CHILD-TUNING, which updates a subset of parameters (called child network) of large pretrained models via strategically masking out the gradients of the non-child network during the backward process.

\section{Reversed Back Propagation -- The Third Pass}
\label{sec:abp}

The third pass is based on the first and the second pass, we assume you have learned those preliminaries, refer to appendix \ref{two passes}. There are a variety of feed-forward or multi-layered neural network, we will introduce our work through a multi-layer perceptron.


Let $\ell$ be the error function of the actual and desired output, $\boldsymbol{y}$ and $\boldsymbol{\hat y}$, of MLP, $u_i^l$ be the total inputs to the neuron $i$ in the $l$-th layer, $v_i^l$ be the output after applying sigmoid activation, and $w_{ij}^l$ be the synaptic weight from neuron $i$ of the $(l-1)$-th to neuron $j$ of the $l$-th layer. We let $v_i^0$ represent the input neurons and $\boldsymbol{x}$ represent the inputs to MLP, $\Delta*$ and $\nabla*$ are the gradients of the second and the third pass respectively. Then we have $\Delta\boldsymbol x = \Delta\boldsymbol v^0 = -{\partial\ell(\boldsymbol{y}, \boldsymbol{\hat y})}/{\partial\boldsymbol v^0} = (\delta_1^0; ...; \delta_{m}^0)$, $m$ is the number of input neurons, $\delta_i^0$ is $-{\partial\ell(\boldsymbol{y}, \boldsymbol{\hat y})}/{\partial v_i^0}$. We also have $\delta_i^l=-{\partial\ell(\boldsymbol{y}, \boldsymbol{\hat y})}/{\partial u_i^l}$ for $l>0$.  Theoretically, $\boldsymbol x$ should be updated by $\Delta\boldsymbol x$ in a gradient descent manner, otherwise, the error of the input neuron would accumulate as the error back-propagation learning process progresses. If $\Delta\boldsymbol x = \boldsymbol0$, then each back-propagation contributes nothing to the accumulated error. The third pass of the net aims to make $\Delta\boldsymbol x$ as close to zeros as possible. We define the input loss $\ell_x$ as

\begin{equation}  \label{third:input loss}
	\ell_x(\Delta\boldsymbol x)=\frac{({\delta_1^0})^2 + ({\delta_2^0})^2 + ... + ({\delta_{m}^0})^2}{2}
\end{equation}

$\ell_x$ formulates how much of the error of input neurons is for a single case. It's expected to be as close to zero as possible. The bigger the $\ell_x$, the larger the accumulated error/gradient/derivative to the input layer. During iterating over many cases, the error accumulated to the input layer growth correspondingly. So $\ell_x$ is the core of our work.

\begin{figure}[h]
	\centering
	\includegraphics[width=0.8\textwidth]{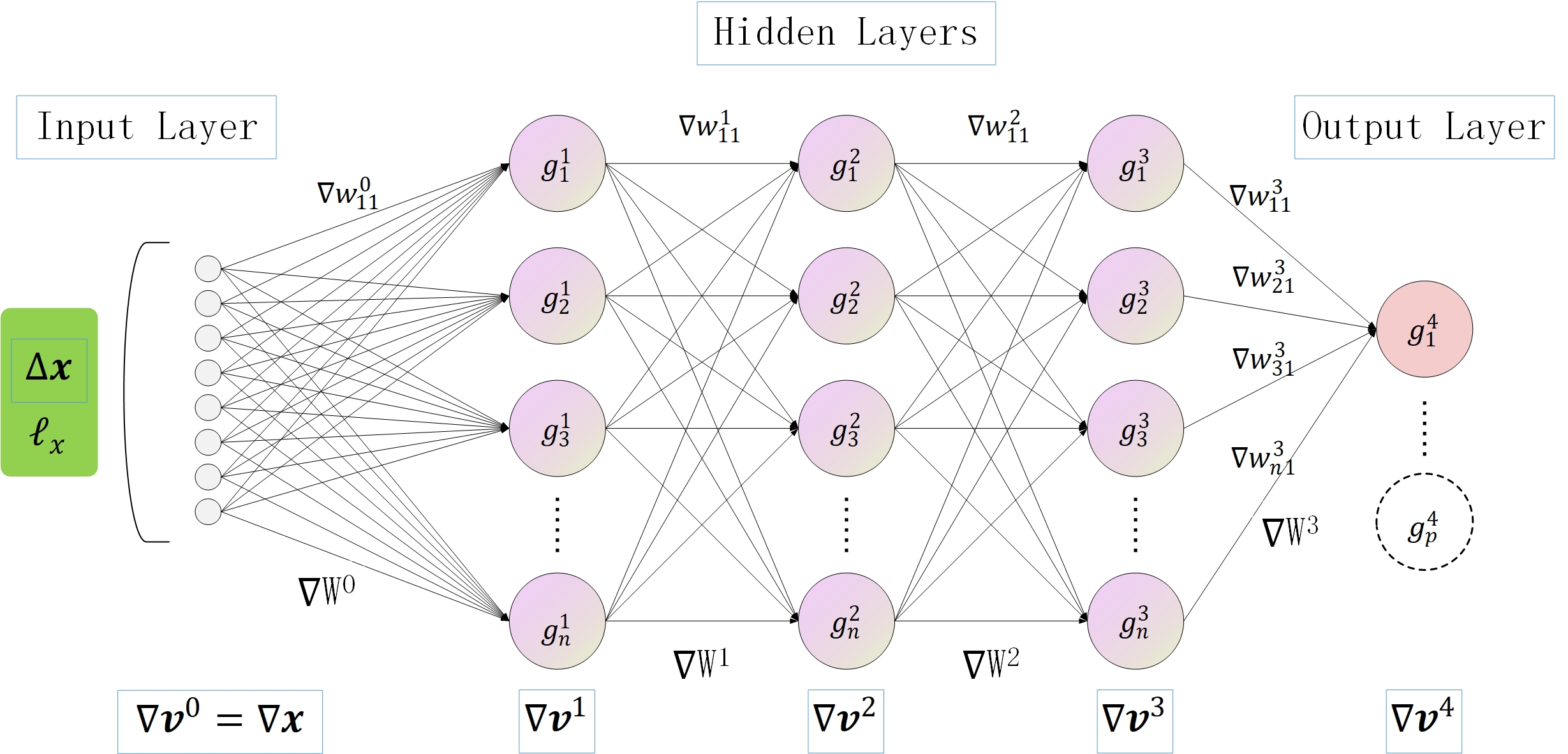}
	\caption{The third pass of a multi-layer perceptron with three hidden layers and corresponding symbols. No bias.}
	\label{another back propagation}
\end{figure}

\paragraph{The third pass}of MLP starts by computing ${\partial \ell_x(\Delta\boldsymbol x)}/{\partial\Delta x}$ for each of the input neurons. Differentiating equation \ref{third:input loss} for a particular input neuron gives

\begin{equation} \label{third:output error}
	\nabla x_i = -\frac{\partial \ell_x(\Delta\boldsymbol x)}{\partial\Delta x_i} = -\delta_i^0
\end{equation}

This time, we need to forward the error signal from input layer to output layer. Analogously, apply the chain rule to get

\begin{align*}
	g_j^0 & = -\delta_j^0		&	\nabla w_{ij}^0 &= g_i^0\delta_j^1, i\ge 1 \\
	g_j^1 &= v_j^1(1-v_j^1)\sum_{i=1}^{m}g_i^0w_{ij}^0
\end{align*}
where $g_i^l$ is $-\partial \ell_x / \partial(-\partial\ell/\partial v_i^l)$, $\nabla w_{ij}^l$ is $-\partial \ell_x / \partial w_{ij}^l$. Apparently, the biases, $w_{0j}^l$, and its gradient, $\Delta w_{0j}^l$, contribute nothing to $\ell_x$ in the second pass, the third pass only consider those weights $w_{i,j}^l$, where $i > 0$.

More generally, we have

\begin{align*}
	\nabla w_{ij}^l &= g_i^l\delta_j^{l+1}, i>0\land l\geq0	&	g_j^l &= v_j^l(1-v_j^l)\sum_{i}g_i^{l-1}w_{ij}^{l-1}, l>0
\end{align*}

Analogously, $\nabla\boldsymbol u^{l+1} = W^l\nabla\boldsymbol v^l, \nabla\boldsymbol v^{l+1} = \boldsymbol v^{l+1}(1-\boldsymbol v^{l+1})\nabla\boldsymbol u^{l+1}, \nabla W^l = \Delta\boldsymbol u^{l+1}(\nabla\boldsymbol v^l)^T$. Repeat these operations until the error signal originating from the input layer propagates to the output layer. Thus, the third pass of the MLP over.

The third pass is the same as the first pass to some extent, except that the biases are not considered no matter whether it exists in the first pass. More details refer to section \ref{implementation:convolution}.

Finally, given a set of input-output pairs, $\{(\boldsymbol x_1, \boldsymbol y_1), ..., (\boldsymbol x_m, \boldsymbol y_m)\}$, the target minimizing both $\ell$ and $\ell_x$ for an optimal $\boldsymbol w$ is

\begin{equation} \label{optim target}
	\boldsymbol w^* = \mathop{\arg\min}_{\boldsymbol w}\sum_{i=1}^{m}\alpha\cdot\ell(\boldsymbol w(\boldsymbol x_i),\boldsymbol y_i)+(1-\alpha)\cdot\ell_x(\frac{\partial\ell(\boldsymbol w(\boldsymbol x_i),\boldsymbol y_i)}{\partial\boldsymbol x_i})
\end{equation}
where $\boldsymbol w$ represents the weights of the neural network, $\alpha$ is a factor ranging from $0$ to $1$ to balance the significance of the gradient in the second pass and the third pass. When setting factor $\alpha$ to $1$, then the target is the same as the other deep learning tasks minimizing the loss function $\ell$.

The procedure of training a neural network via three passes is outlined in Algorithm \ref{alg:three passes}. In fact, as described in the algorithm, the target in equation \ref{optim target} cannot be optimized by one pass. It's necessary to separate the target into two terms, one is $\ell$ and the other is $\ell_x$. With three passes learning, the neural network may be able to escape from the local minima as illustrated in figure \ref{error surface}, but there is no guarantee, it may perform worse. 

Vanishing gradient problem is common in deep learning, as the neural networks go deeper, it is more likely to observe such phenomena. In the third pass of the net, it remains. However, the second and the third pass are in the opposite direction, and their gradient vanishes in the opposite direction too, the gradients of the two passes can compensate for each other. We believe that with proper factor $\alpha$ to balance the gradients in the two passes, the vanishing gradient problem can be properly resolved.


\noindent\begin{minipage}{.5\textwidth}
\begin{algorithm}[H]
	\caption{Training a neural network via three passes}
	\label{alg:three passes}
	\begin{algorithmic}[1]
		\State {\bfseries Input:} a set of data $\{(\boldsymbol x_1, \boldsymbol y_1), ..., (\boldsymbol x_m, \boldsymbol y_m)\}$, factor $\alpha$, epoch $n$, thresh $\epsilon$, learning rate $\eta$, neural network $\boldsymbol w$
		\State {\bfseries Output:} the trained neural network $\boldsymbol w^*$, total error $E$, total input error/loss $E_x$
		\State	initialize $\boldsymbol w$
		\State	$epoch\gets 0$
		\Repeat
		\State	$E, E_x, \Delta\boldsymbol w, \nabla\boldsymbol w\gets0, 0, \boldsymbol0, \boldsymbol0$
		\For{$i\gets1$ to $m$}
		\State $l=\ell(\boldsymbol w(\boldsymbol x_i),\boldsymbol y_i)$	\Comment{First Pass}
		\State  $\Delta\boldsymbol w\gets\Delta\boldsymbol w+\partial l/\partial\boldsymbol w$	\Comment{Second Pass}
		\State $l_x=\ell_x(\partial l/\partial\boldsymbol x_i)$	\Comment{Second Pass}
		\State $\nabla\boldsymbol w\gets\nabla\boldsymbol w+{\partial l_x}/{\partial\boldsymbol w}$	\Comment{Third Pass}
		\State	$E\gets E+l$
		\State	$E_x\gets E_x+l_x$
		\EndFor
		\State $\boldsymbol w\gets\boldsymbol w+\eta(\alpha$$\Delta\boldsymbol w$ + (1 - $\alpha$)$\nabla\boldsymbol w$)
		\State $epoch\gets epoch + 1$
		\Until{$epoch<n$ \textbf{or} $E\leq\epsilon$}
	\end{algorithmic}
\end{algorithm}
\end{minipage}
\begin{minipage}[]{.5\textwidth}
	\centering
	\includegraphics[width=0.8\textwidth]{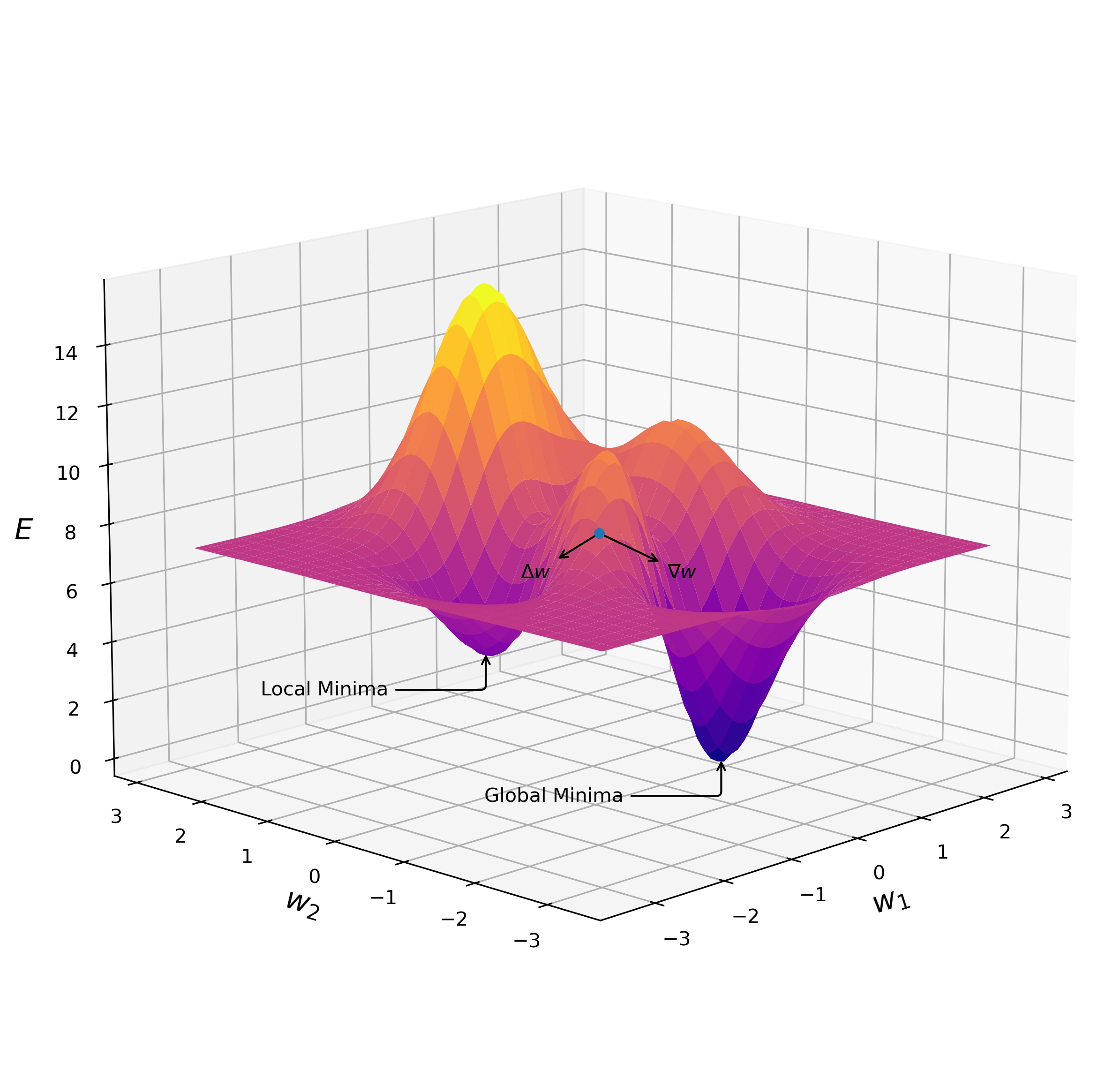}
	\captionof{figure}{Considering the error surface as a function of the neural network input instead of its weight, then a new problem arises, and transforming the gradient of the input($\Delta\boldsymbol x$), line 10 of algorithm \ref{alg:three passes}, to the gradient of the weight($\nabla\boldsymbol w$) would make the gradient descent($\Delta\boldsymbol w$) more interesting.}\label{error surface}
\end{minipage}

\section{Implementation of the third pass of convolution}
\label{implementation:convolution}

Here, we demonstrate in detail the third pass of the convolution operation in figure \ref{abp_overview} and algorithm \ref{alg:three passes}. The second pass is the neural network back-propagation pass which is contrary to the forward pass, and the third pass is contrary to the back-propagation passes, we called it another back-propagation, ABP, for clarity.

The convolution forward with kernel size $ks$, stride $1$, input channel $C_{in}$, output channel $C_{out}$, no bias is

\begin{equation} \label{abpi:eq1}
	\boldsymbol y(i,j,k) = \sum_{c=0}^{C_{in}-1}\sum_{h=0}^{ks-1}\sum_{w=0}^{ks-1}\boldsymbol w(i, c, h, w)\boldsymbol x(c, j+h, k+w)
\end{equation}

where
$\boldsymbol x\in\mathbf{R}^{C_{in}\times(H+ks-1)\times(W+ks-1)}$ is input,
$\boldsymbol y\in\mathbf{R}^{C_{out}\times H\times W}$ is output,
$\boldsymbol w\in\mathbf{R}^{C_{out}\times C_{in}\times ks\times ks}$ is parameter.

Be aware of that the back propagation algorithm computes the gradient along the inverse direction of the forward propagation layer-wisely, i.e., reverse those equations like equation~\ref{abpi:eq1}. Then, the back propagation of convolutions is

\begin{equation} \label{abpi:eq2}
	\nabla\boldsymbol w(i, j, k, l)= \sum_{h=0}^{H-1}\sum_{w=0}^{W-1}\boldsymbol x(j, k+h, l+w)\nabla\boldsymbol y(i, h, w)
\end{equation}

\begin{equation}\label{abpi:eq3}
	\begin{split}
		\nabla\boldsymbol x(i,j,k)=\sum_{c=0}^{C_{out}-1}\sum_{h=0}^{ks-1}\sum_{w=0}^{ks-1}\nabla\boldsymbol y_0(c, j+h, k+w)\boldsymbol w(c, i, ks-h-1,ks-w-1)
	\end{split}
\end{equation}

where $\nabla\boldsymbol y_0$ is $\nabla\boldsymbol y$'s 0-padding of size $ks-1$ on each border.
$\nabla\boldsymbol w$ is the gradient of convolution parameters used for updating the parameters itself, $\nabla\boldsymbol x$ would propagate to former layer to compute their gradients. For those parameter free operations, such as pooling or activate functions, there is no $\nabla\boldsymbol w$, that means we only need to conduct some equations like~\ref{abpi:eq3} to get $\nabla\boldsymbol x$ and propagate it to former layer.

As for ABP, it regards the back propagation process as the forward propagation process, because ABP aims at minimizing $\ell_x$ which is a function of the model parameters and its gradient gained in the (first) back propagation process. ABP adopts absolutely the same mechanism as the back propagation algorithm. ABP computes the gradient along the inverse direction of the (first) back propagation layer-wisely, i.e., reverse those equations like equation~\ref{abpi:eq3}. In fact, equation~\ref{abpi:eq3} is similar to equation~\ref{abpi:eq1}, both are the convolution forward operation, the former just flips the kernel of the latter.

Consequently, ABP is formulated as

\begin{equation} \label{abpi:eq4}
	\begin{split}
		\Delta\boldsymbol w(i, j, ks-k-1, ks-l-1) = \sum_{h=0}^{H+ks-2}\sum_{w=0}^{W+ks-2}&\nabla\boldsymbol y_0(i, k+h, l+w)\Delta\boldsymbol x(j, h, w)
	\end{split}
\end{equation}

\begin{equation} \label{abpi:eq5}
	\Delta\boldsymbol y_0(i,j,k) = \sum_{c=0}^{C_{in}-1}\sum_{h=0}^{ks-1}\sum_{w=0}^{ks-1}\boldsymbol w(i, c, h, w)\Delta\boldsymbol x_0(c, j+h, k+w)
\end{equation}

where $\Delta\boldsymbol *$ represent the gradient in ABP, $\Delta\boldsymbol x_0$ is $\Delta\boldsymbol x$'s 0-padding of size $ks-1$ on each border. For those parameter free operations, such as pooling or activate functions, there is no $\Delta\boldsymbol w$, that means we only need to conduct those equations like equation~\ref{abpi:eq5} to get $\Delta\boldsymbol y_0$ and propagate it to ``former'' layer.

The ABP is an inverse process of the backward propagation, but it is not identical to the forward propagation in figure~\ref{abp overview}. The bias in convolution contributes nothing to $\nabla\boldsymbol x$ in the backward propagation(though we assume there is no bias), i.e., $\nabla\boldsymbol x$ showing in equation ~\ref{abpi:eq3} is a function irrelevant to bias. So, ABP concentrates on the parameters weights, but ignores the bias.

It seems there is no shortcut \footnote{The deep learning framework provides a finite set of operations such as convolution and its backward, if you need some novel operations, the framework may not well support, implementing it in CUDA is the only way, or that would make you desperate no matter how many CPUs you have.} to compute equation~\ref{abpi:eq4} except following the formula naively. But equation~\ref{abpi:eq2} and equation~\ref{abpi:eq4} adopt the same operation essentially. Refer to figure \ref{abp overview}, $\nabla W$ and $\Delta W$ are the twins and seem to could be calculated in the same mechanism, because both have similar input. We can repeat an operation identical to equation~\ref{abpi:eq2} and then flip the last 2 dimensions of the results to achieve the same effect as equation~\ref{abpi:eq4}.

Moreover, equation~\ref{abpi:eq5} is identical to the convolution forward in equation~\ref{abpi:eq1} absolutely. So, for those parameter free operations, at least most of them, conducting ABP on them is just conducting their forward propagation operation again.

Thus, we completely implement ABP for convolution operation smoothly without any new operations which may need to be implemented from scratch. Besides, the convolution forward and its backpropagation provided by the frameworks are accelerated by special algorithms, for example, fast Fourier transform would accelerate those convolutions with large kernel size. Therefore, transforming ABP operations to the forward or backward operations could take the advantage of it. 

Note that only the case where the convolution stride of $1$ is considered here. As for stride greater than $1$, we re-implemented the second and the third pass of convolution based on unfold and fold operation - im2col algorithm. It is very slow compared to what we have introduced above, so we only use it when necessary.

\section{Weight Initialization}
\label{sec:weight init}

\cite{bradley2010learning} found that back-propagation gradients were smaller as one moves from the output layer towards the input layer, just after initialization. He studied networks with linear activation at each layer, finding that the variance of the back-propagated gradients decreases as we go backwards in the network. Based on the smaller and smaller variance as backward propagation progresses, \cite{KaimingHe2015} expects the variance of the responses or back-propagated gradients at each layer to be the same, \cite{glorot2010understanding} maintains both activation variances and back-propagated gradients variance as one moves up or down the network. The main difference is the former addresses rectifier non-linearity, and the latter considers the symmetric activation with unit derivative at $0$.

In the three passes learning, the actual gradient descent is

\begin{equation} \label{init:eq1}
\boldsymbol w=\boldsymbol w+\eta(\alpha\Delta\boldsymbol w + (1 - \alpha)\nabla\boldsymbol w)
\end{equation}

It's different from standard weight initialization method, which keeps the variance of response or gradient of each layer the same. Here, the actual gradient consists of $\Delta\boldsymbol w$ and $\nabla\boldsymbol w$. So we need to keep the sum of the gradient variance over the two passes in each layer the same to avoid gradient diminishing or explosion. Rectifier non-linearity and symmetric activation with unit derivative at $0$ are studied on the $d$-layer perceptron separably.
\paragraph{Symmetric activation with unit derivative at $0$}
Consider the hypothesis that we are in a linear regime at the initialization, that the weights are initialized independently and that the input features variances are the same ($=Var[\boldsymbol x]$). Then we can say that, with $n_i$ the size of layer $i$ and $\boldsymbol x$ the network input, then we have $f^\prime(u^i_k)\approx1$, and
\begin{equation} \label{linear: eq1}
	Var[\boldsymbol v^i]=Var[\boldsymbol x]\prod_{i^\prime=0}^{i-1}n_{i^\prime}Var[W^{i^\prime}]
\end{equation}
\begin{equation} \label{linear: eq2}
	Var[\Delta\boldsymbol v^i]=Var[\Delta\boldsymbol v^d]\prod_{i^\prime=i}^{d-1}n_{i^\prime+1}Var[W^{i^\prime}]
\end{equation}
\begin{equation} \label{linear: eq3}
	Var[\nabla\boldsymbol v^i]=Var[\nabla\boldsymbol v^0]\prod_{i^\prime=0}^{i-1}n_{i^\prime}Var[W^{i^\prime}]
\end{equation}
Note $\boldsymbol v^0=\boldsymbol x$ here. And according to equation \ref{third:output error}, $Var[\nabla\boldsymbol v^0]=Var[\Delta\boldsymbol v^0]$, substitute it to equation \ref{linear: eq3}, then we get
\begin{equation} \label{linear: eq4}
	Var[\Delta\boldsymbol v^i+\nabla\boldsymbol v^i]=Var[\Delta\boldsymbol v^i]+Var[\Delta\boldsymbol v^0]\prod_{i^\prime=0}^{i-1}n_{i^\prime}Var[W^{i^\prime}]
\end{equation}
\paragraph{ReLU activation} Note $v_i^l=max(0,u_i^l)$, and we introduce the prior that $W^l$ has a symmetric distribution around zero, and assume that the activation derivative w.r.t. $u_i^l$ and $\Delta v_i^l$ are independent of each other. Then get $Var[\boldsymbol u^l]=n_{l-1}Var[W^{l-1}\boldsymbol v^{l-1}]=n_{l-1}Var[W^{l-1}]E[(\boldsymbol v^{l-1})^2]$, $E[(\boldsymbol v^l)^2] = \frac{1}{2}Var[\boldsymbol u^l]$ and 
\begin{equation} \label{rectifier: eq1}
	Var[\boldsymbol u^i]=n_0Var[\boldsymbol x]Var[W^0]\prod_{i^\prime=1}^{i-1}\frac{1}{2}n_{i^\prime}Var[W^{i^\prime}],\quad\forall i>0
\end{equation}
\begin{equation} \label{rectifier: eq2}
	Var[\Delta\boldsymbol v^i]=Var[\Delta\boldsymbol v^d]\prod_{i^\prime=i}^{d-1}\frac{1}{2}n_{i^\prime+1}Var[W^{i^\prime}],\quad\forall i\ge0
\end{equation}
In the third pass, we also have $\nabla\boldsymbol v^l=f^\prime(\boldsymbol u^l)\nabla\boldsymbol u^l$. For the ReLU case, $f^\prime(u_i^l)$ is zero or one, and their probabilities are equal. Assume $f^\prime(\boldsymbol u^l)$ and $\nabla\boldsymbol u^l$ are independent of each other(it seems absurd, but not), then $E[\nabla\boldsymbol v^l] = \frac{1}{2}E[\nabla\boldsymbol u^l]=0$, and also $E[(\nabla\boldsymbol v^l)^2]=Var[\nabla\boldsymbol v^l]=\frac{1}{2}Var[\nabla\boldsymbol u^l]$. Then we have
\begin{equation} \label{rectifier: eq3}
	Var[\nabla\boldsymbol v^i]=Var[\nabla\boldsymbol v^0]\prod_{i^\prime=0}^{i-1}\frac{1}{2}n_{i^\prime}Var[W^{i^\prime}],\quad\forall i\ge0
\end{equation}
substitute $Var[\nabla\boldsymbol v^0]=Var[\Delta\boldsymbol v^0]$ to equation \ref{rectifier: eq3},
\begin{equation} \label{rectifier: eq4}
	Var[\Delta\boldsymbol v^i+\nabla\boldsymbol v^i]=Var[\Delta\boldsymbol v^i]+Var[\Delta\boldsymbol v^0]\prod_{i^\prime=0}^{i-1}\frac{1}{2}n_{i^\prime}Var[W^{i^\prime}]
\end{equation}
The equations \ref{linear: eq1} \ref{linear: eq2} and \ref{rectifier: eq1} \ref{rectifier: eq2} are identical to the equations in \cite{glorot2010understanding} and \cite{KaimingHe2015} respectively, please refer to the original paper for more details.
Let equation \ref{linear: eq4} or \ref{rectifier: eq4} be the same for $\forall i$, i.e. $Var[\Delta\boldsymbol v^i+\nabla\boldsymbol v^i]=Var[\Delta\boldsymbol v^{i+1}+\nabla\boldsymbol v^{i+1}]$, get
\begin{equation} \label{init1}
	n_{i+1}Var[W^i]-1=n_{i+1}Var[W^i](n_iVar[W^i]-1)\prod_{i^\prime=0}^{i-1}n_{i^\prime}n_{i^\prime+1}Var[W^{i^\prime}]^2
\end{equation}
\begin{equation} \label{init2}
	\frac{1}{2}n_{i+1}Var[W^i]-1=\frac{1}{2}n_{i+1}Var[W^i](\frac{1}{2}n_iVar[W^i]-1)\prod_{i^\prime=0}^{i-1}\frac{1}{4}n_{i^\prime}n_{i^\prime+1}Var[W^{i^\prime}]^2
\end{equation}
both come to the similar results, that is $n_iVar[W^i]=n_{i+1}Var[W^i]=1$ or $n_iVar[W^i]=n_{i+1}Var[W^i]=2$. And the conclusion is also the same as original paper. In \cite{KaimingHe2015}, they declare that $n_iVar[W^i]=2$ or $n_{i+1}Var[W^i]=2$ works fine, but in our case of the three passes learning, both of equations are satisfied is necessary.
The equation \ref{init1} and \ref{init2} could be satisfied if the neural network has the same width, $n_i$, on each layer. However, we can get the conclusion without the condition of $Var[\nabla\boldsymbol v^0]=Var[\Delta\boldsymbol v^0]$, there may be some other feasible solutions in equation \ref{init1} and \ref{init2}, which are derived from the unused conditions. For example, $Var[W^i]=1/\sqrt{n_in_{i+1}}$ and $n_{i+1}=n_i$ can also make the equation \ref{init1} hold, and $Var[W^i]=2/\sqrt{n_in_{i+1}}$ and $n_{i+1}=n_i$ can still make the equation \ref{init2} hold. If we suppress the condition of $n_{i+1}=n_i$, our reformulated initialization approach is $Var[W^i]=1/\sqrt{n_in_{i+1}}$ or $Var[W^i]=2/\sqrt{n_in_{i+1}}$. Our conclusion could be approximately fulfilled if the majority layer of a neural network has the same width, and the gradient in the second and the third pass contribute equal, i.e., $\alpha=0.5$.

We think those equations have no essential distinction compared to the original paper. And it's hard to find some circumstances where our initialization formula performs better. For instance, the convolution neural network nowadays normally consists of several stages, and each stage has the same or similar width, then $n_{i+1}=n_i$ in that stage, and our $Var[W^i]=2/\sqrt{n_in_{i+1}}$ degrade to $Var[W^i]=2/n_i$ which is identical to \cite{KaimingHe2015}, so our method actually has no superiority.


\section{Experiments}
\label{sec:experiment}

We re-implement the second pass and the third pass of the neural network in Pytorch \cite{NEURIPS2019_9015}, evaluate our method on the MNIST, CIFAR10, and CIFAR100 dataset. Considering the conclusion of our weight initialization strategy is similar to \cite{KaimingHe2015}, we adopt its weight initialization strategy rather ourselves'. Its counterparts are trained in two passes learning, which is the standard back-propagation algorithm. Optimization is performed using SGD with momentum 0.9, a mini-batch size of 128, and no weight decay. The default splitting of the MNIST and CIFAR datasets is adopted. We want to explore the effects of the three passes training, so there is no batch normalization, residual connections, and weight decay, a completely pure neural network only with ReLU activation was what we want. And we don’t exactly follow Algorithm \ref{alg:three passes} to perform the three passes learning, that is the trained model is updated every batch rather every epoch. All the training metric we report here is averaged over an epoch.
\paragraph{Plain Networks}Plain-20 and plain-32 in table \ref{tab:net} are designed for CIFAR-10 and CIFAR-100, its architecture is analogous to ResNet \cite{he2016deep}, except that there are no residual connections and batch normalization. We think it's enough to verify the effects of our three passes learning. All the experiments of CIFAR are conducted on them.

\begin{table}[h]
	\centering
	\begin{tabular}{lc|cc}
		\hline
		\multicolumn{1}{l|}{layer name} & output size   & \multicolumn{1}{c|}{plain-20}                 & plain-32                  \\ \hline
		\multicolumn{1}{l|}{conv1}      & $32\times 32$ & \multicolumn{2}{c}{$3\times 3, 16$}                                       \\ \hline
		\multicolumn{1}{l|}{conv2\_x}   & $32\times 32$ & \multicolumn{1}{c|}{$[3\times 3,16]\times 6$} & $[3\times 3,16]\times 10$ \\ \hline
		\multicolumn{1}{l|}{conv3\_x} &
		$16\times 16$ &
		\multicolumn{1}{c|}{\begin{tabular}[c]{@{}c@{}}$3\times 3,32$, stride $2$\\ $[3\times 3,32]\times 5$\end{tabular}} &
		\begin{tabular}[c]{@{}c@{}}$3\times 3, 32$, stride $2$\\ $[3\times 3,32]\times 9$\end{tabular} \\ \hline
		\multicolumn{1}{l|}{conv3\_x} &
		$8\times 8$ &
		\multicolumn{1}{c|}{\begin{tabular}[c]{@{}c@{}}$3\times 3,64$, stride $2$\\ $[3\times 3,64]\times 5$\end{tabular}} &
		\begin{tabular}[c]{@{}c@{}}$3\times 3, 64$, stride $2$\\ $[3\times 3,64]\times 9$\end{tabular} \\ \hline
		\multicolumn{1}{l|}{}           & $1\times 1$   & \multicolumn{2}{c}{avg pool,$\{10,100\}$-d fc, softmax}                       \\ \hline
		\multicolumn{2}{c|}{Params(M)}                  & \multicolumn{1}{c|}{0.27}                     & 0.46                      \\ \hline
	\end{tabular}
	\caption{Both nets have no batch normalization and residual connections, each 3x3-Conv followed by a ReLU activate function.}
	\label{tab:net}
\end{table}

\subsection{Weight Initialization}


We trained convolution network plain-20 and plain-32 to investigate the effects of weight initialization. When they are trained under a learning rate of 0.1, only our approach converges, both Kaiming \cite{KaimingHe2015} and Xavier \cite{glorot2010understanding} stall. In fact, performing three passes training with learning rate $\eta=0.1$, $\alpha= 0.1$ is equivalent to $\eta_1=0.01$ for gradient of the second pass and $\eta_2=0.09$ for gradient of the third pass according to equation \ref{init:eq1}, so it is in some sense lead to inconsistent experimental conditions. Then we further set $\eta=0.01$ for the other initialization methods for a fair comparison. Table \ref{tab:weight init} confirms that our method makes the neural network learn better, plain-20 always gets higher training error and lower validation error compared to its counterparts. Our three passes learning initializes the model with \cite{KaimingHe2015}, and all methods don't apply weight decay. We further trained deeper neural networks, but all approaches stalled, and the gradient vanished in the input layer. And we attempt to initialize all layers of a model with the same uniform distribution or normal distribution but fail in the end. Therefore, weight initialization is still a significant procedure of neural net training.
\begin{table}[h]
	\centering
	\begin{tabular}{|cc|cc|cc|}
		\hline
		\multicolumn{2}{|c|}{\multirow{2}{*}{}}                                       & \multicolumn{2}{c|}{CIFAR-10}      & \multicolumn{2}{c|}{CIFAR-100}     \\ \cline{3-6} 
		\multicolumn{2}{|c|}{}                                                        & \multicolumn{1}{c|}{train} & val   & \multicolumn{1}{c|}{train} & val   \\ \hline
		\multicolumn{1}{|c|}{\multirow{3}{*}{plain-20}} & Xavier\cite{glorot2010understanding} & \multicolumn{1}{c|}{$95.66$} & $85.83$ & \multicolumn{1}{c|}{$62.62$} & $55.16$ \\ \cline{2-6} 
		\multicolumn{1}{|c|}{}                          & Kaiming\cite{KaimingHe2015} & \multicolumn{1}{c|}{$95.31$} & $86.10$ & \multicolumn{1}{c|}{$62.82$} & $55.17$ \\ \cline{2-6} 
		\multicolumn{1}{|c|}{}                          & $\alpha=0.1$                & \multicolumn{1}{c|}{$92.48$} & $87.29$ & \multicolumn{1}{c|}{$61.98$} & $57.17$ \\ \hline
		\multicolumn{1}{|c|}{\multirow{3}{*}{plain-32}} & Xavier                      & \multicolumn{1}{c|}{$90.31$} & $82.87$ & \multicolumn{1}{c|}{$52.58$} & $46.79$ \\ \cline{2-6} 
		\multicolumn{1}{|c|}{}                          & Kaiming                     & \multicolumn{1}{c|}{$90.85$} & $83.14$ & \multicolumn{1}{c|}{$50.59$} & $45.26$ \\ \cline{2-6} 
		\multicolumn{1}{|c|}{}                          & $\alpha=0.1$                & \multicolumn{1}{c|}{$91.68$} & $85.88$ & \multicolumn{1}{c|}{$57.92$} & $54.59$ \\ \hline
	\end{tabular}
	\caption{Both plain-20 and plain-32 are trained for 200 epochs and we report their best training and validation accuracy on CIFAR. The plain nets initialized by Xavier or Kaiming are trained in two passes training, that is the vanilla back-propagation. The three passes training is performed with $\alpha$ of $0.1$.}
	\label{tab:weight init}
\end{table}

We have observed the degradation problem - the 32-layer plain net has higher training error throughout the whole training procedure, even though the solution space of the 20-layer plain network is a subspace of that of the 32-layer one. This is consistent with the results in \cite{he2016deep}. And our approach dampened thus degradation to a large extent.

\begin{figure}[h]
	\centering
	\begin{subfigure}{.3\textwidth}
		\centering
		\includegraphics[width=\textwidth]{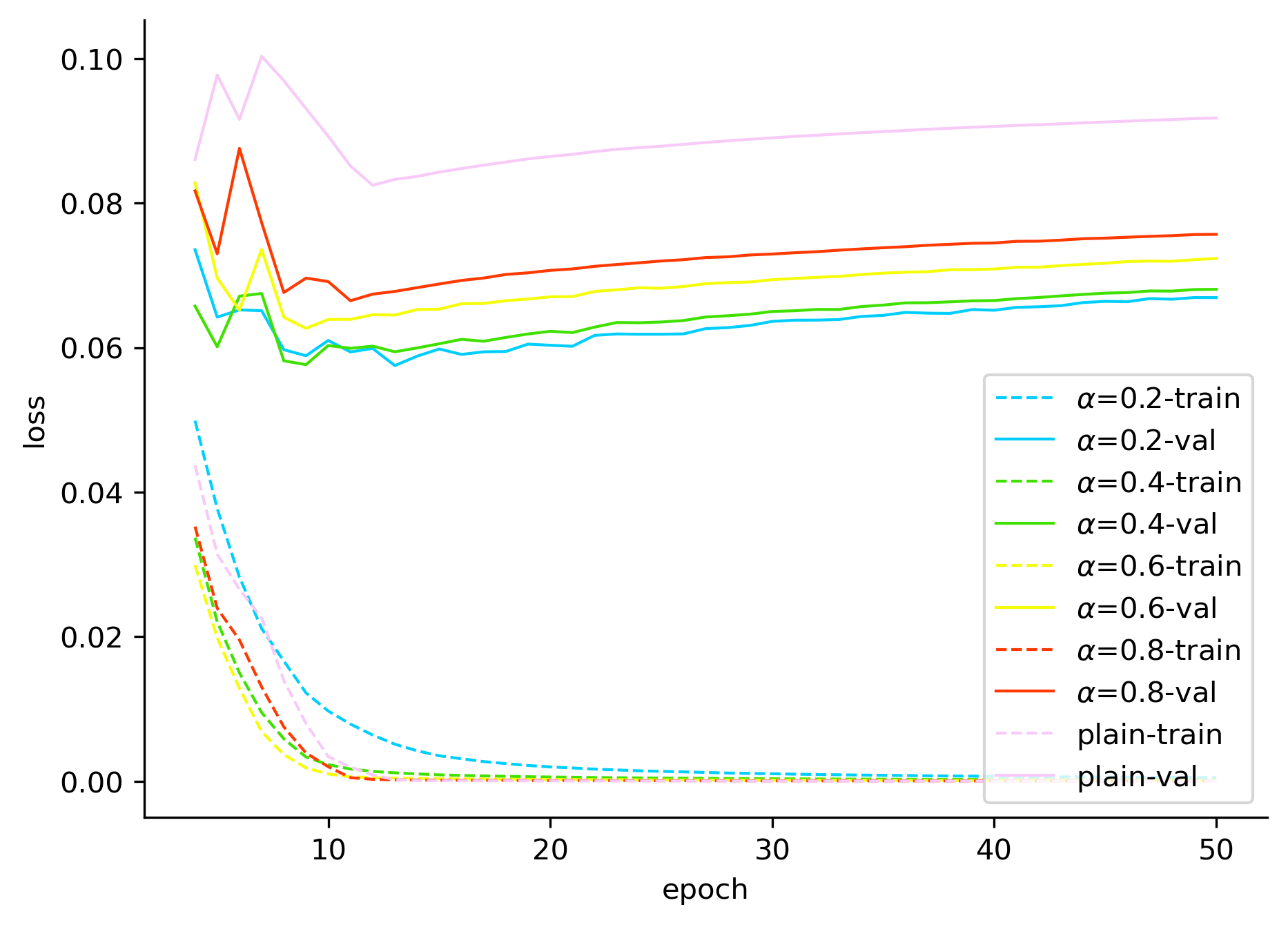}
	\end{subfigure}
	\begin{subfigure}{.3\textwidth}
		\centering
		\includegraphics[width=\textwidth]{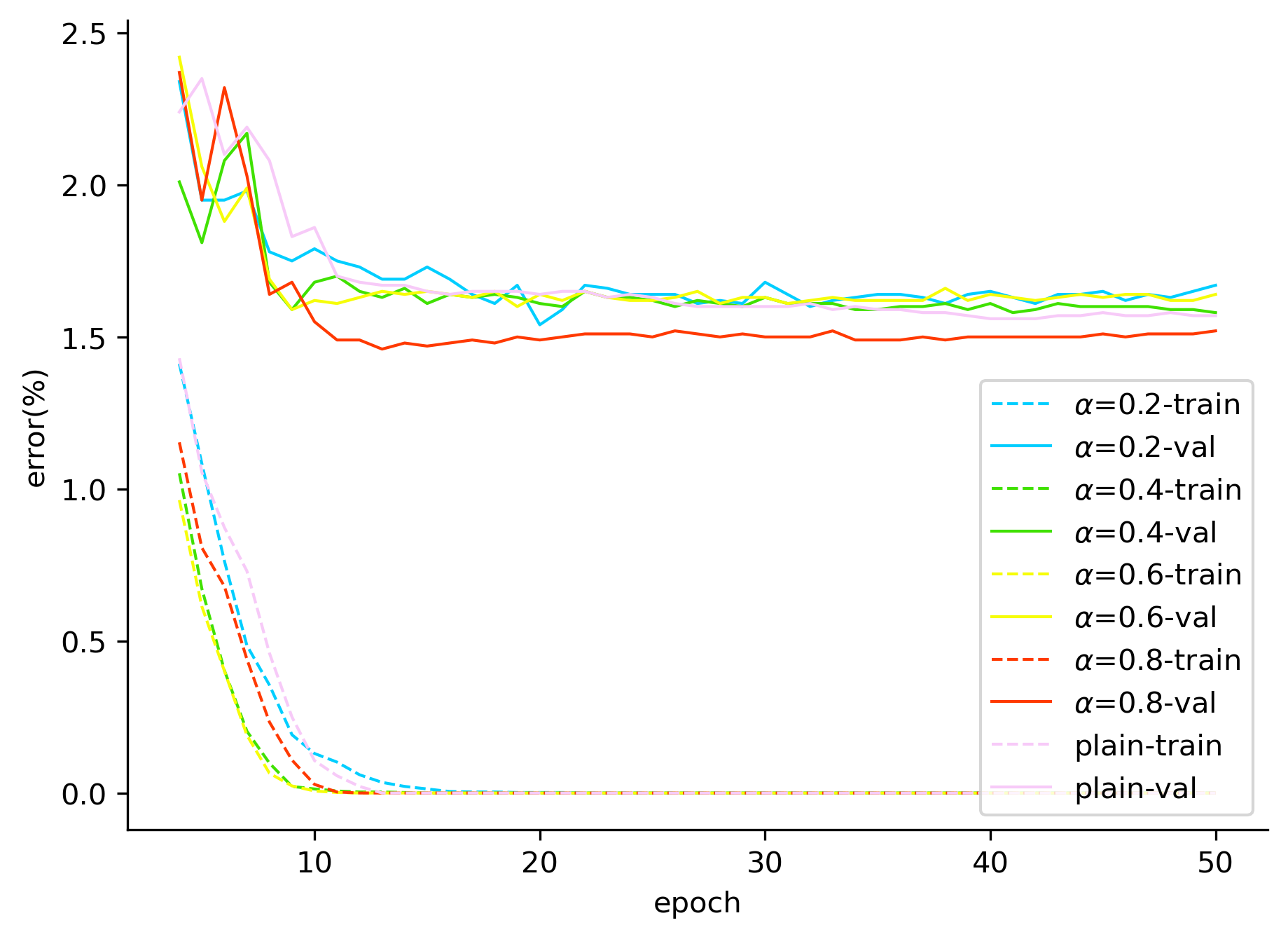}  
	\end{subfigure}
	\begin{subfigure}{.3\textwidth}
		\centering
		\includegraphics[width=\textwidth]{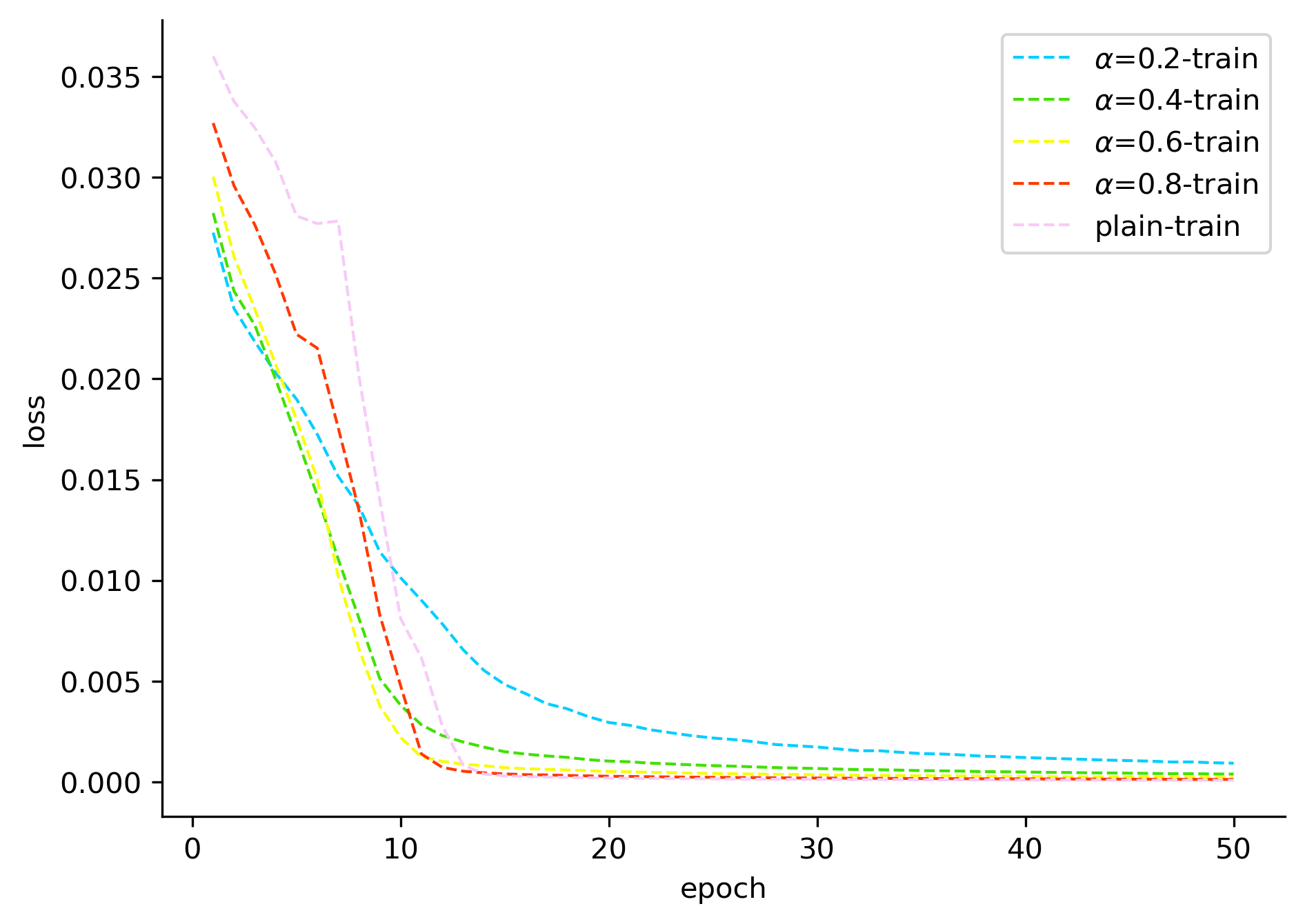}  
	\end{subfigure}
	\caption{\textbf{Left:} MNIST train and validation loss, $\ell$ in equation \ref{sedond:loss func}. \textbf{Middle:} train and validation error. \textbf{Right:} input loss, $\ell_x$ in equation \ref{third:input loss}. Plain represents the neural net trained in vanilla back-propagation, and others are its counterparts applying three passes learning with different $\alpha$ respectively.}
	\label{exp: mnist}
\end{figure}

\subsection{MNIST and Analysis}
\label{mnist analysys}

There is no data augmentation for MNIST, we just flatten the image and feed it to a 3-layer perceptron with 784 neurons in the input layer, 512 neurons in the hidden layer, 10 neurons in the output layer. The learning rate is 0.1, and the 3-layer perceptron is trained for 50 epochs from scratch.


\paragraph{Unexpected phenomenon}In figure \ref{exp: mnist}, the input loss gets small as $\alpha$ gets large, this is a strange phenomenon because the input loss should be proportional to $\alpha$. The small $\alpha$ implies we expect the $\ell_x$ term in equation ~\ref{optim target} to be constrained heavier, thus smaller input loss. And the input loss doesn't accumulate but continues to decrease in any cases which is contrary to our hypothesis. We suspect the dataset is so simple that it can easily be overfitted by the MLP, and the overfitted MLP is well converged on the training dataset, it almost has no gradient to update itself, so it has a small input loss even without the constraint of the $\ell_x$ term in equation ~\ref{optim target}. And the small $\alpha$ constrained the MLP to not overfit , thus making the training loss converge slow, then further getting small validation loss. Therefore, the smaller the $\alpha$, the slower the input loss in figure \ref{exp: mnist} will converge, further leading to the bigger input loss.

The smaller the $\alpha$, the smaller the valid loss. But the valid loss does not seem to correspond to the generalization error here. $\alpha=0.8$ has a higher validation loss than that $\alpha$ of lower than $0.8$, but it has the lowest validation error. In fact, there is no obvious accuracy gap among all of those curves.

We further studied the case where $\alpha=0$, refer to Appendix \ref{alpha0}.

\begin{figure}[h]
	\centering
	\begin{subfigure}{.28\textwidth}
		\centering
		\includegraphics[width=\textwidth]{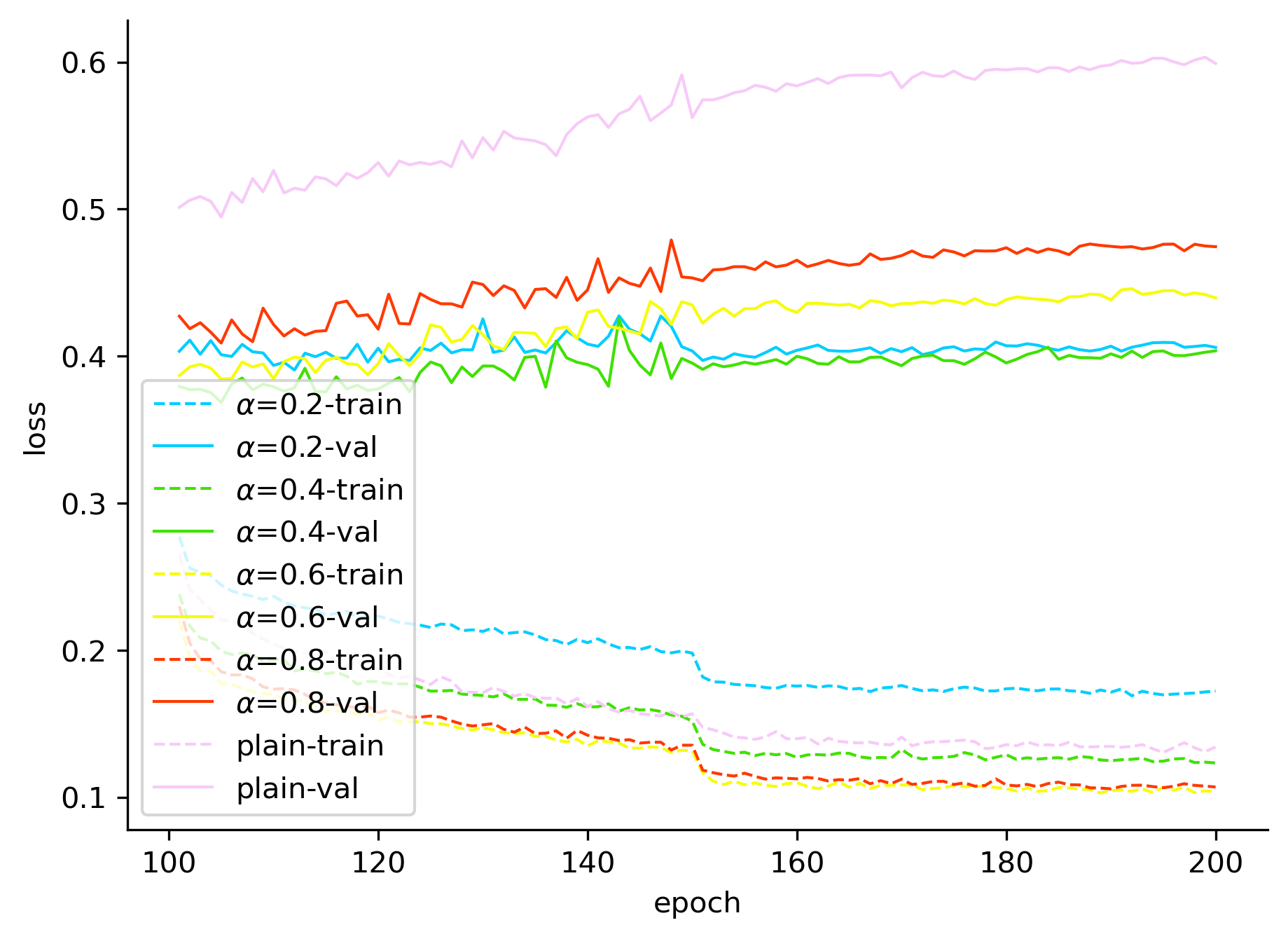}
	\end{subfigure}
	\begin{subfigure}{.28\textwidth}
		\centering
		\includegraphics[width=\textwidth]{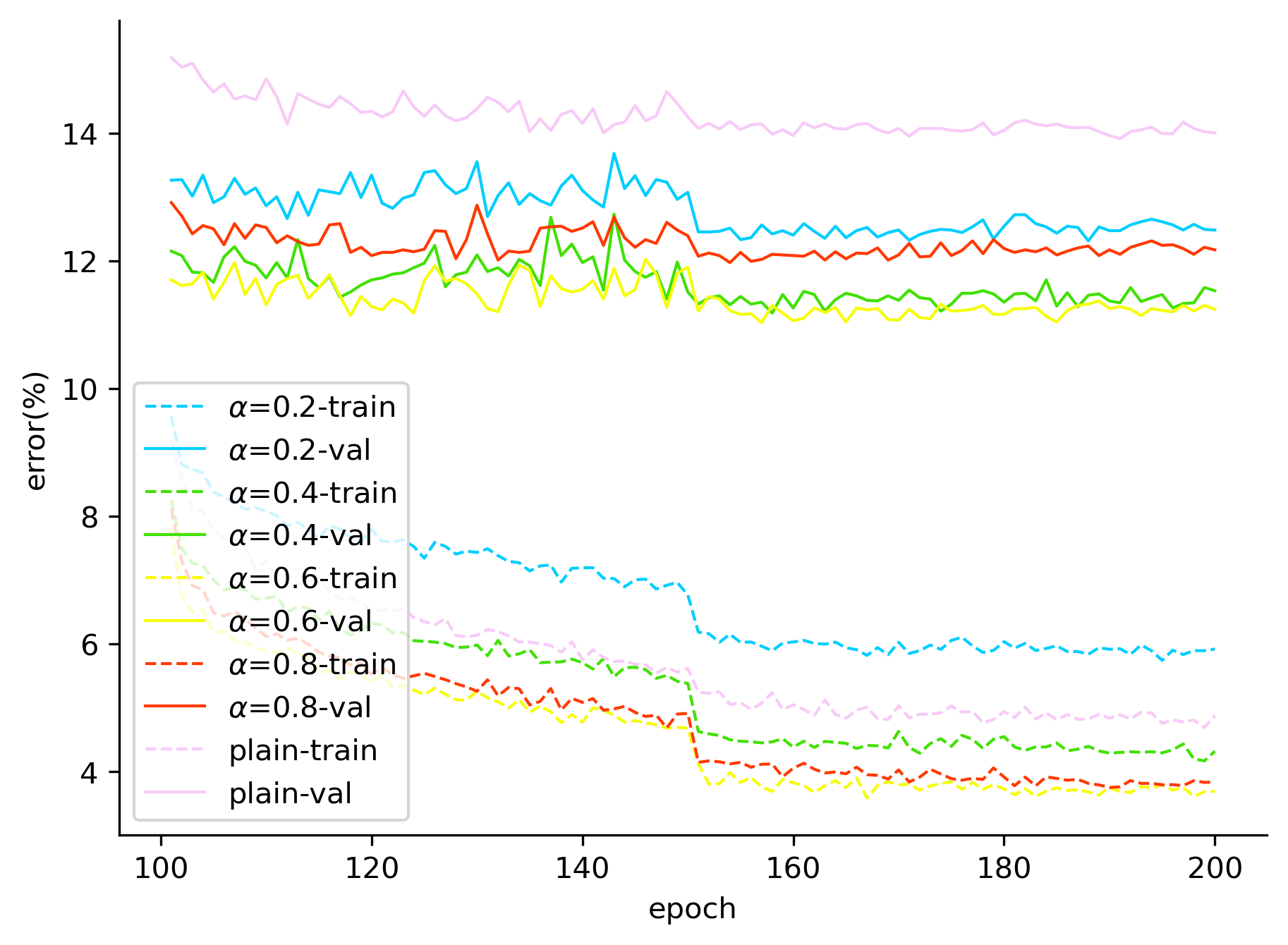}  
	\end{subfigure}
	\begin{subfigure}{.28\textwidth}
		\centering
		\includegraphics[width=\textwidth]{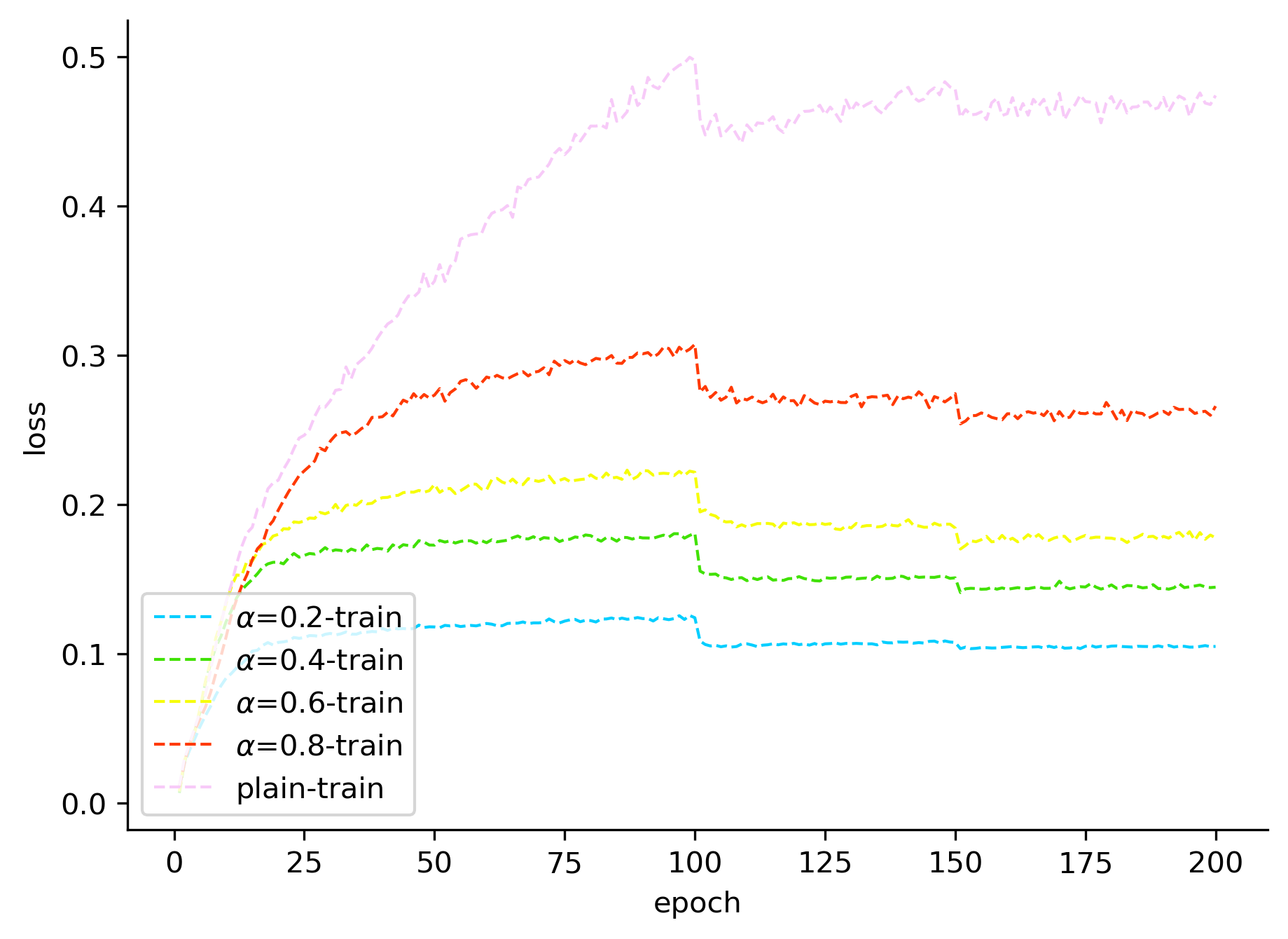}  
	\end{subfigure}
	
	\begin{subfigure}{.28\textwidth}
		\centering
		\includegraphics[width=\textwidth]{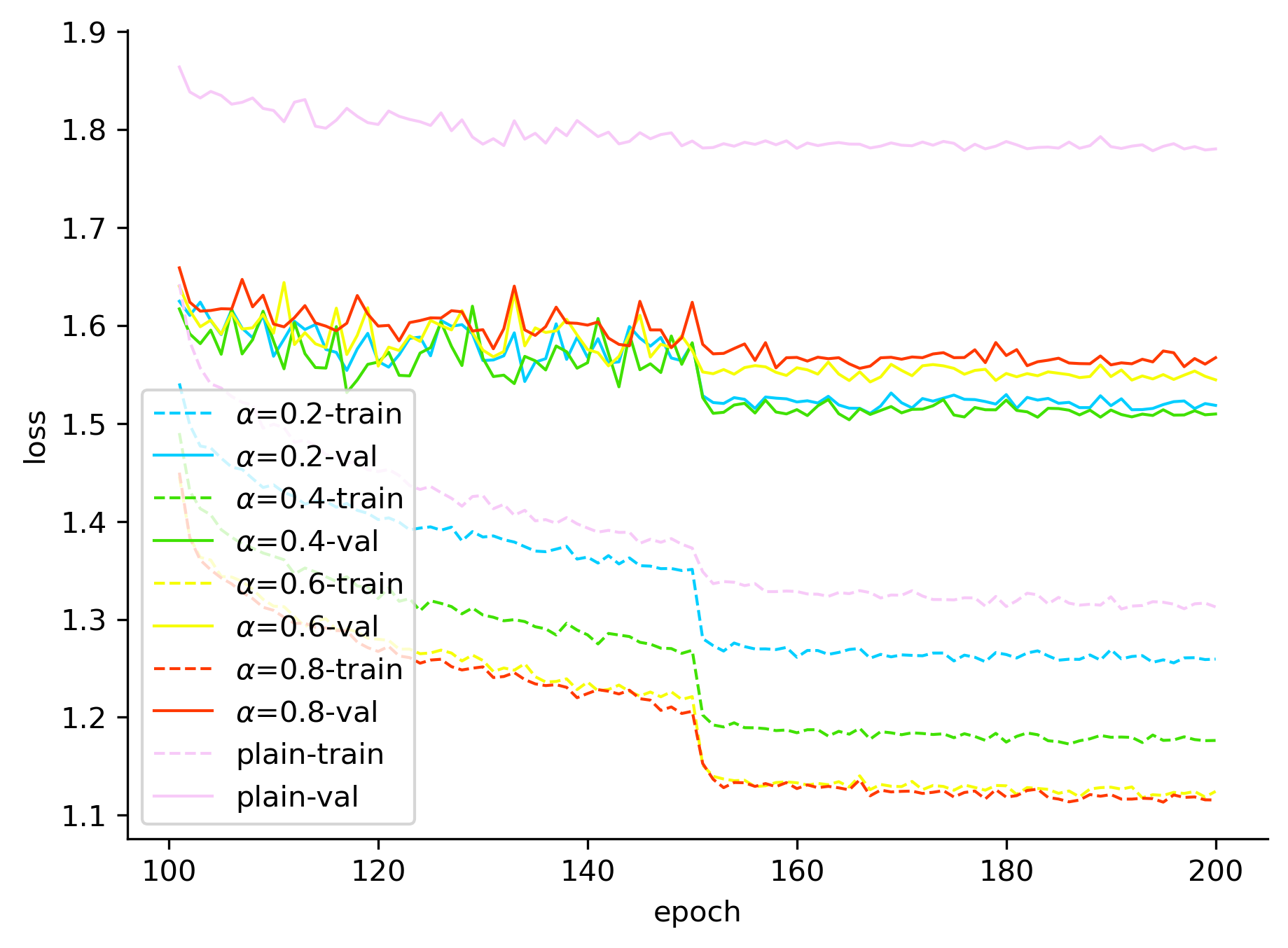}
		\caption{Train and validation loss.}
	\end{subfigure}
	\begin{subfigure}{.28\textwidth}
		\centering
		\includegraphics[width=\textwidth]{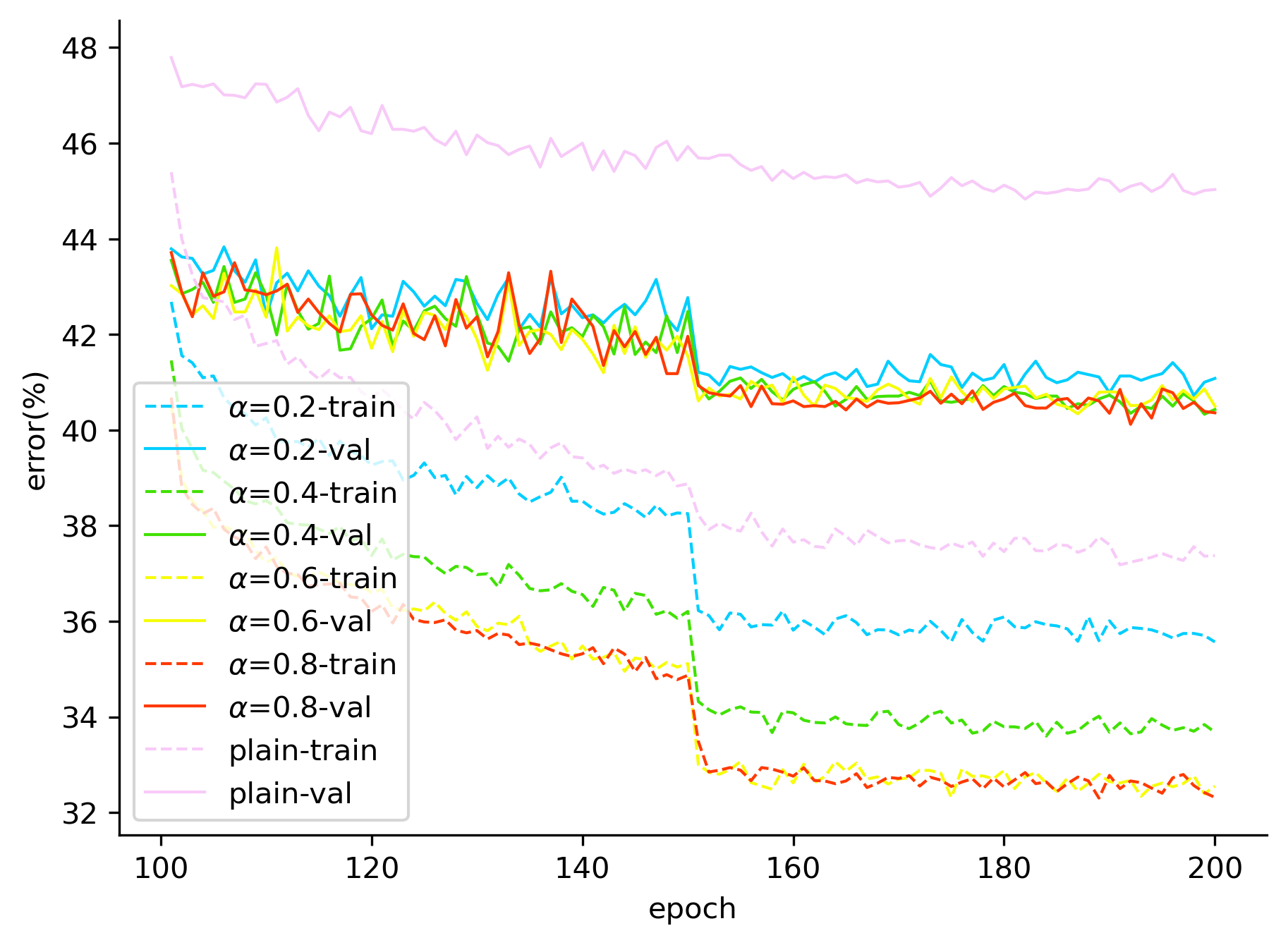}
		\caption{Train and validation error.}
	\end{subfigure}
	\begin{subfigure}{.28\textwidth}
		\centering
		\includegraphics[width=\textwidth]{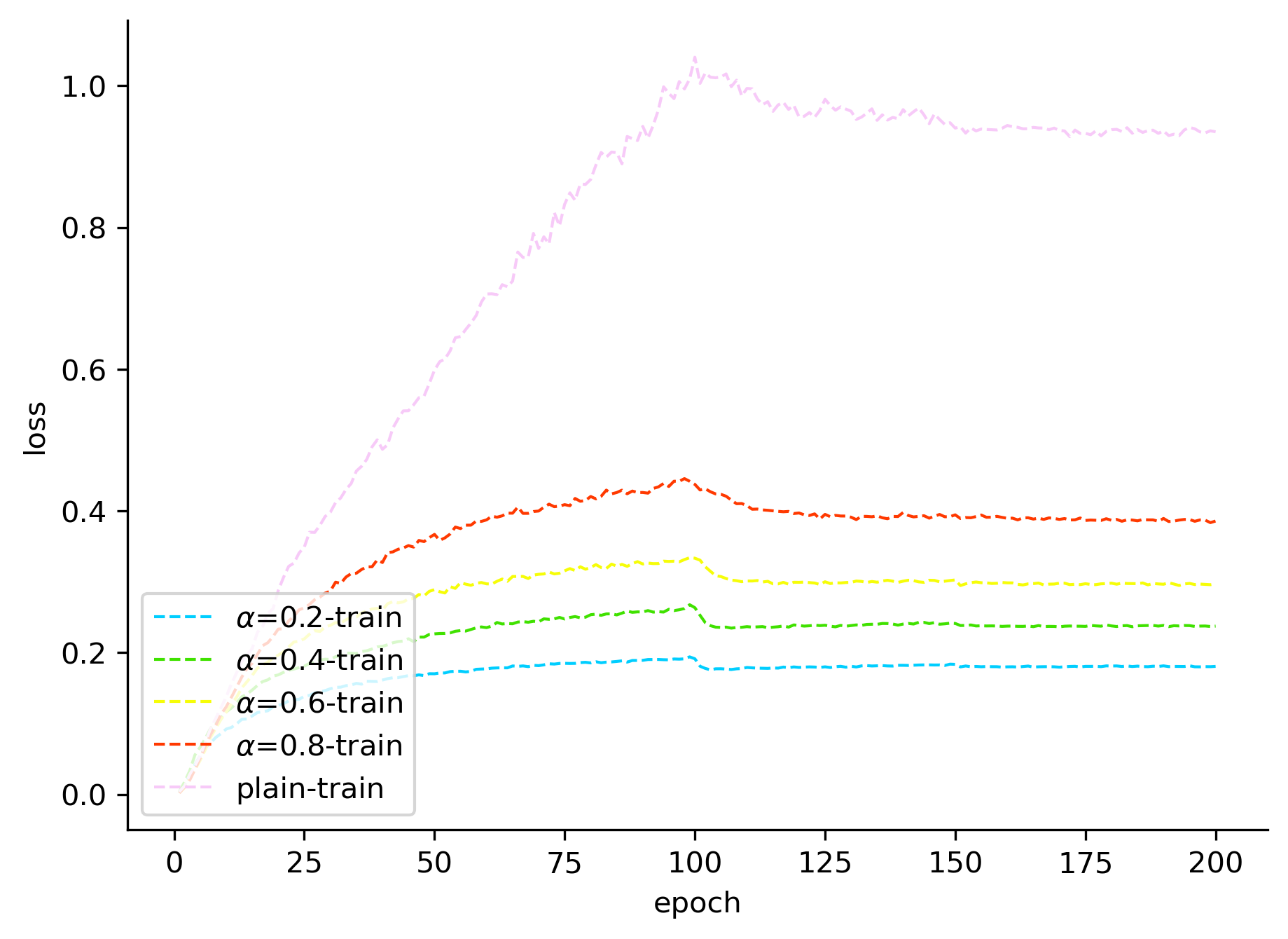}
		\caption{Input loss.}
	\end{subfigure}
	\caption{\textbf{Top:} Training on CIFAR-10. \textbf{Bottom:} Training on CIFAR-100. Plain-20 and its counterparts applying three passes learning.}
	\label{exp:cifar}
\end{figure}

\subsection{CIFAR and Analysis}

During training, images of CIFAR are randomly horizontally flipped and zero-padded on each side with four pixels before taking a random 32 × 32 crop. Mean and standard deviation normalization is also applied. The convolutional neural network we trained here is plain-20. It is trained for 200 epochs from scratch with an initial learning rate of 0.01 and decay by a factor of 10 at 100 and 150 epoch. In the three passes training, the learning rate is adjusted according to $\alpha$ for a fair comparison, that is ${\eta}/{\alpha}$. Intuitively, the initial learning rate should be larger at the beginning, and then gradually decreases as the model converges, but we found the model without batch normalization and residual connections and weight decay does not fit such a big learning rate.


\paragraph{Three passes learning Versus Two passes learning}Plain nets are trained under two passes learning and three passes learning is performed with $\alpha=0.2,0.4,0.6,0.8$. Figure \ref{exp:cifar} seems different from figure \ref{exp: mnist}, the input loss gets small as $\alpha$ gets small, the performance of omitted cases of $\alpha=0.1,0.3,0.5,0.7,0.9$ is analogous to $\alpha=0.2,0.4,0.6,0.8$, and the input loss accumulates as the training process progresses in all those cases. All our conjectures or intentions have been fulfilled here. The partial derivative of the input layer is optimized by the third pass successfully, and the net has persistent gain correspondingly. Two passes learning has higher input loss than three passes learning due to the error accumulated to the input layer neurons has been ignored by it, and three passes learning adds one more pass to optimize the error accumulated to the input layer neurons, thus getting smaller input loss. $\alpha=0.2$ gets a higher training error but a smaller validation error compared to plain net on CIFAR-10. We think this is in some sense reflect that three passes learning could make the neural network get rid of many local minima while the two passes learning may be stuck at local minima, thus it's more likely to get a lower training error but a higher validation error. Generally, we refer to such phenomenon as overfitting. And we also observed the underfitting on CIFAR-100.

The case of $\alpha=0.2$ gets the lowest input loss, but neither does the lowest input loss mean the lowest validation error, nor the highest input loss. There is a balance between the input loss and the validation error, and so does the gradient in the second pass and the third pass. 

Similar settings conducted on plain-32 produce a consistent effect. Our method is even better. However, we tried using the sigmoid activate function to train a deep neural network but failed in the end. It seems the gradient vanishing problem does not meet our expectations, we can't compensate the gradient of the second pass via the third pass because the gradient vanished long before it flows into the input layer when the network is equipped with a sigmoid activate function, the error accumulated to the input layer is almost zero, which further makes it unnecessary to perform the third pass, details refer to \cite{glorot2010understanding}.

\section{Conclusion}

Training with three passes is less sensitive to hyper-parameters, for instance, learning rate, batch size. Many neural networks work bad on too small batch size or inappropriate learning rate, but our approaches perform well (robust enough) on the same circumstance. Besides, $\alpha=0$ deserves our attention, because it completely abandoned the traditional back-propagation process and take effects in some extent. And conduct the third pass may double the training time but have no influence at inference time. Recently, finetuning a Transformer model on downstream tasks emerges in various fields, our approach may fit such circumstances.

The third pass possesses innegligible potential, many aspects of it have yet to be explored, such as the fourth pass or further, adjusting factor $\alpha$ dynamically, fine grained $\alpha$ searching, separable optimizers for the second and the third pass, and so on, and experiments on it are still not adequate. All in all, it deserves studying, there still are some unexplainable phenomena, and remain some nuts to be addressed.



\bibliographystyle{unsrt}  
\bibliography{references}

\appendix

\section{The first and second pass}
\label{two passes}

Figure \ref{feed_forward_network} shows the architectural graph of a MLP. A neuron in any layer of the network is connected to all the nodes/neurons in the previous layer. Signal flow through the network progresses in a forward direction, from left to right on a layer-by-layer basis.

\begin{figure}[h]
	\centering
	\includegraphics[width=0.8\textwidth]{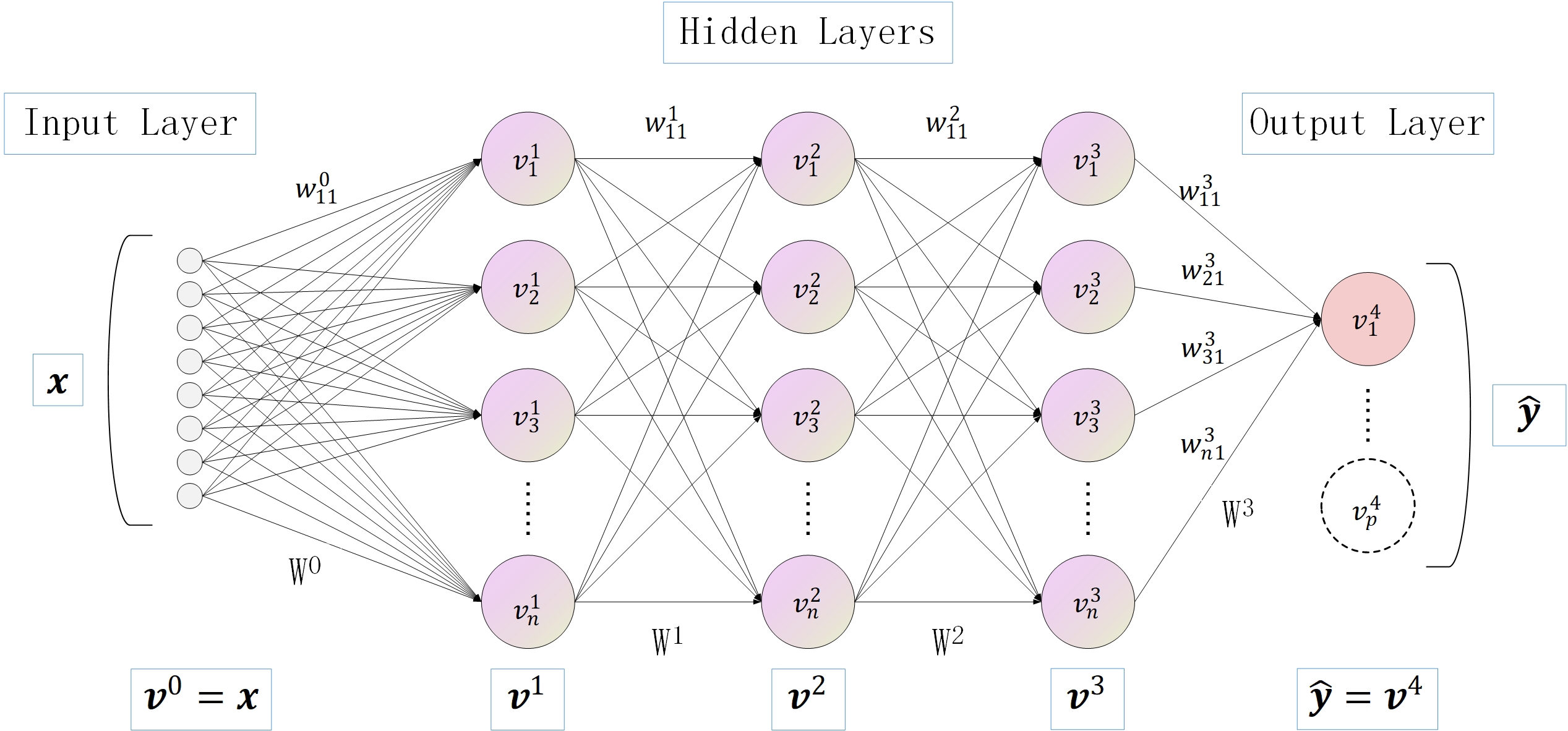}
	\caption{The first pass of a multi-layer perceptron with three hidden layers and corresponding symbols. We omit the biases(threshold) $w_{0j}^l$ for clarity.}
	\label{feed forward network}
\end{figure}

\paragraph{The first pass}The total input, $u^l_j$, to the neuron $j$ in the $l$-th layer is a linear function of the outputs,  $v^{l-1}_i$, of those neurons that are connected to $j$ and of the corresponding synaptic weights, $w_{ij}^{l-1}$, and of the biases, $w_{0j}^{l-1}$

\begin{equation} \label{first:linear sum}
	u_j^l=\sum_{i}w_{ij}^{l-1}v_i^{l-1}+w_{0j}^{l-1}
\end{equation}
where $l \in \mathbf{N^+}$.

The neuron $j$ in the $l$-th layer then apply activate function to transform the linear sum

\begin{equation} \label{first:activate}
	v_j^l=\frac{1}{1+e^{-u_j^l}}
\end{equation}


If we treat bias $w_{0j}^3$ as a synaptic weight driven by a fixed input equal to $+1$ and combine it with weights, $W^3$, to shape $W_0^3\in\mathbf{R}^{p\times(n+1)}$, and add an extra constant unit indexed by $0$ to $\boldsymbol v^3$ to form $\boldsymbol v_0^3\in\mathbf{R}^{n+1}$, then we have $\boldsymbol u^{4}=W_0^{3}\boldsymbol v_0^{3}, \boldsymbol v^{4}=1/(1+exp(-\boldsymbol u^{4}))$. Analogously, repeat the $\boldsymbol u-\boldsymbol v$ transforms until the signal $\boldsymbol x$ from the input layer flows into the output layer. Thus, the first pass of the MLP finished.

If there is a fixed, finite set of input-output cases, the error in the performance of the net with a particular set of weights can be computed by comparing the actual and desired output vectors for every case. The error, $\ell$, for a single case is defined as

\begin{equation} \label{sedond:loss func}
	\ell(\boldsymbol{\hat y}, \boldsymbol y)=\frac{1}{2}\sum_{j}(\hat{y}_{j}-y_{j})^2
\end{equation}
where $j$ is an index over output neurons, $\boldsymbol{\hat{y}}$ is the actual state of output neurons and $\boldsymbol y$ is their desired state. $\boldsymbol{\hat{y}}$ here is $(v_1^4; v_2^4;...; v_p^4)$.

\begin{figure}[h]
	\centering
	\includegraphics[width=0.8\textwidth]{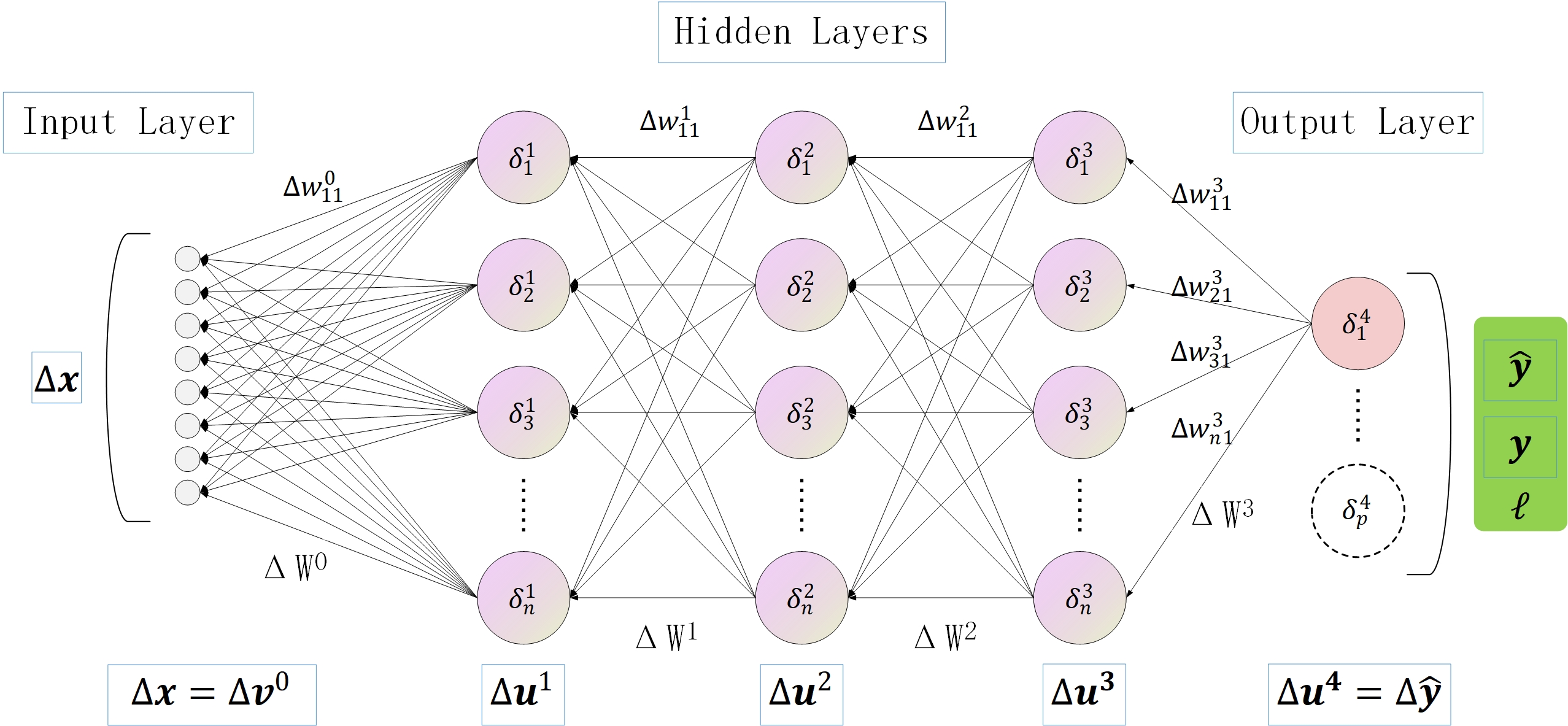}
	\caption{The second pass of a multi-layer perceptron with three hidden layers and corresponding symbols. We omit the gradient of biases $\Delta w_{0j}^l$ for clarity.}
	\label{back propagation}
\end{figure}

\paragraph{The second pass}of MLP starts by computing ${\partial \ell(\boldsymbol{\hat y}, \boldsymbol y)}/{\partial \hat{y}}$ for each of the output neurons. Differentiating equation \ref{sedond:loss func} for a particular neuron gives

\begin{equation} \label{second:output error}
	\Delta\hat y_j = -\frac{\partial \ell(\boldsymbol{\hat y}, \boldsymbol y)}{\partial\hat y_j} = y_j - \hat{y}_j = y_j - v_j^4
\end{equation}

We can then apply the chain rule to get 

\begin{align*}
	\delta_i^4 & = v_j^4(1-v_j^4)(y_j-v_j^4)		&	\Delta w_{ij}^3 &= \delta_j^4v_i^3, i\ge 1 \\
	\delta_i^3 &= v_i^3(1-v_i^3)\sum_{j=1}^{p}\delta_j^4w_{ij}^3	&	\Delta w_{0j}^3 &= \delta_j^4
\end{align*}
where $\delta_i^l$ is $-\partial \ell / \partial u_i^l$ if $l\neq0$ else $-\partial \ell / \partial v_i^l$, $\Delta w_{ij}^l$ is $-\partial \ell / \partial w_{ij}^l$.


More generally, we have

\begin{align*}
	\Delta w_{0j}^l &= \delta_j^{l+1}	&	\Delta w_{ij}^l &= v_i^l\delta_j^{l+1}, i>0	&	\delta_i^l &= (v_i^l(1-v_i^l)\mathbb{1}_{l\neq0} + \mathbb{1}_{l=0})\sum_{j}\delta_j^{l+1}w_{ij}^l
\end{align*}
where $0\leq l<4$ in this case.

Analogously, $\Delta\boldsymbol v^l = (W^l)^T\Delta\boldsymbol u^{l+1}, \Delta\boldsymbol u^l = \boldsymbol v^l(1-\boldsymbol v^l)\Delta\boldsymbol v^l, \Delta W_0^l = \Delta\boldsymbol u^{l+1}(\boldsymbol v_0^l)^T$ for $l < 4$. Repeat these operations until the error signal propagates to the input layer. Thus, the second pass of the MLP finished.

\section{$\alpha = 0$}
\label{alpha0}

\begin{figure}[h]
	\centering
	\begin{subfigure}{.22\textwidth}
		\centering
		\includegraphics[width=\textwidth]{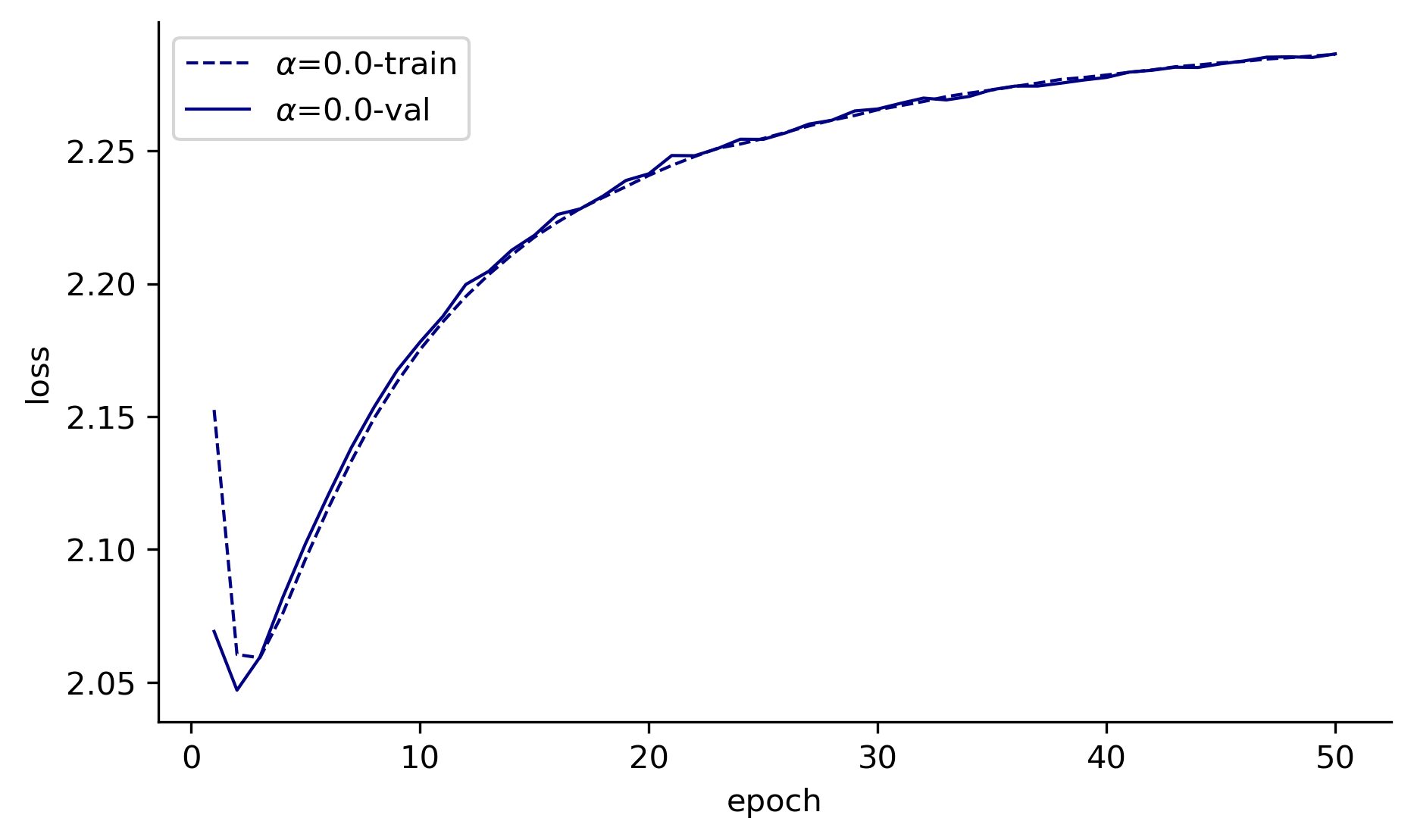}
	\end{subfigure}
	\begin{subfigure}{.22\textwidth}
		\centering
		\includegraphics[width=\textwidth]{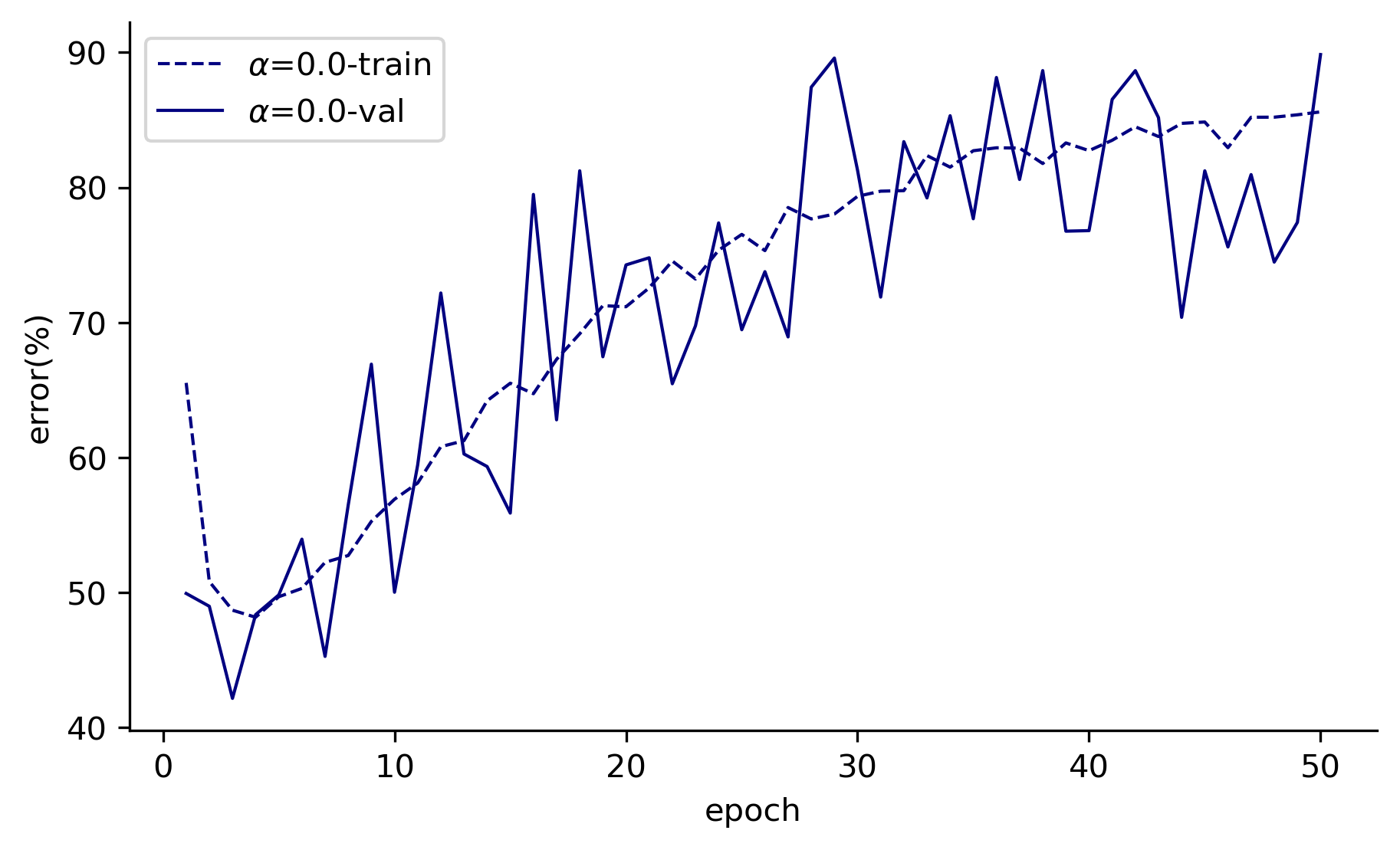}
	\end{subfigure}
	\begin{subfigure}{.22\textwidth}
		\centering
		\includegraphics[width=\textwidth]{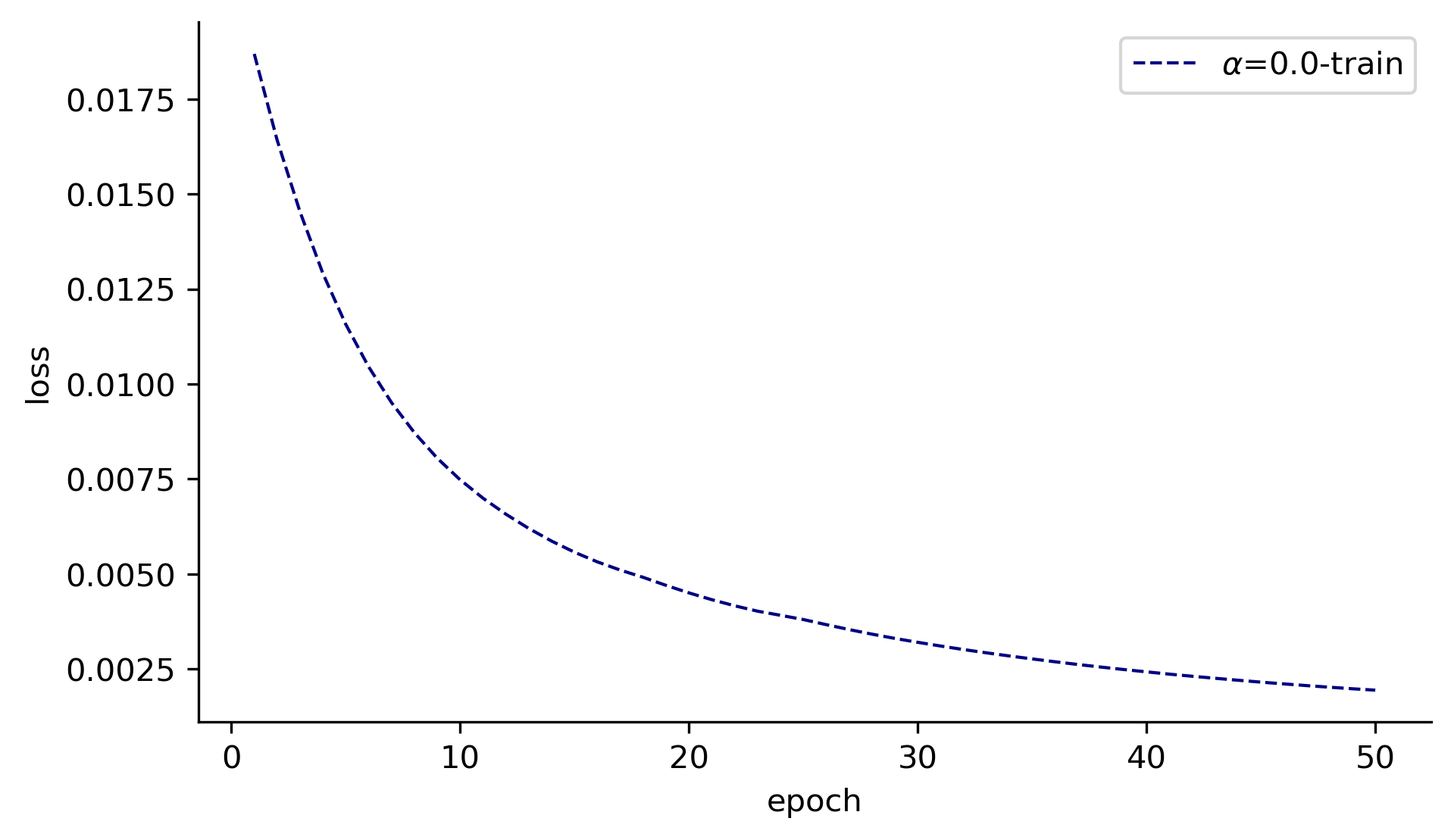}
	\end{subfigure}

	\begin{subfigure}{.22\textwidth}
		\centering
		\includegraphics[width=\textwidth]{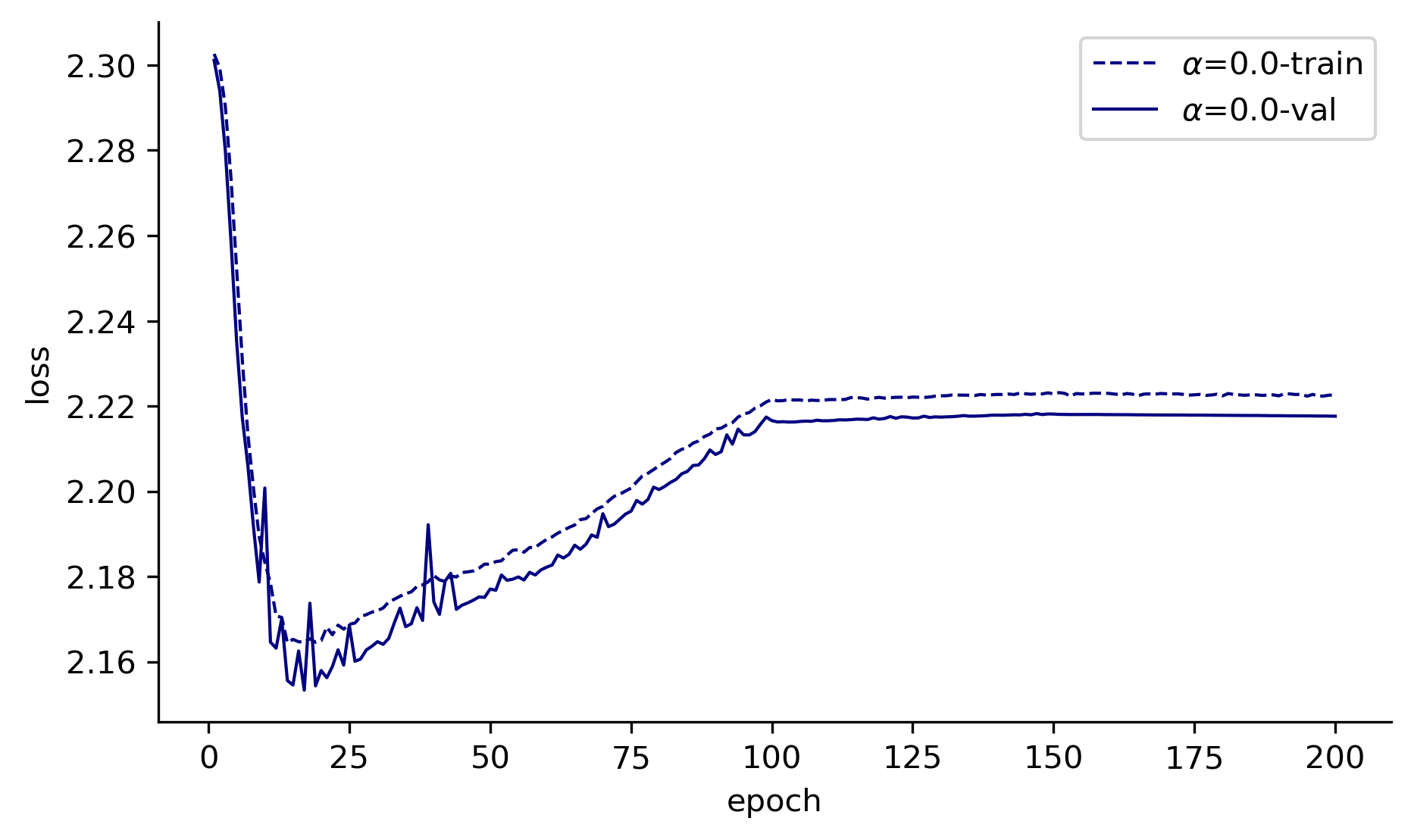}
	\end{subfigure}
	\begin{subfigure}{.22\textwidth}
		\centering
		\includegraphics[width=\textwidth]{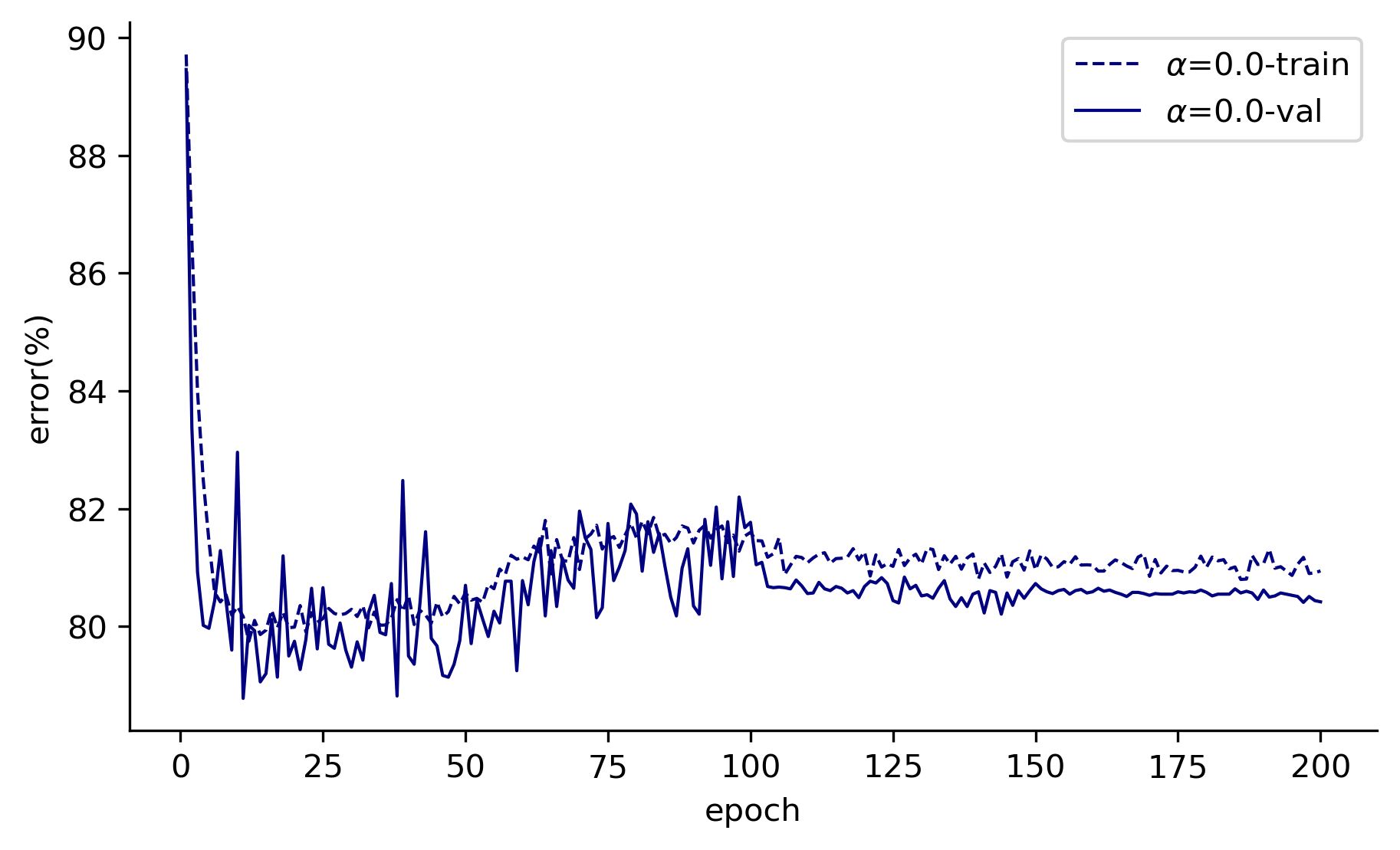}
	\end{subfigure}
	\begin{subfigure}{.22\textwidth}
		\centering
		\includegraphics[width=\textwidth]{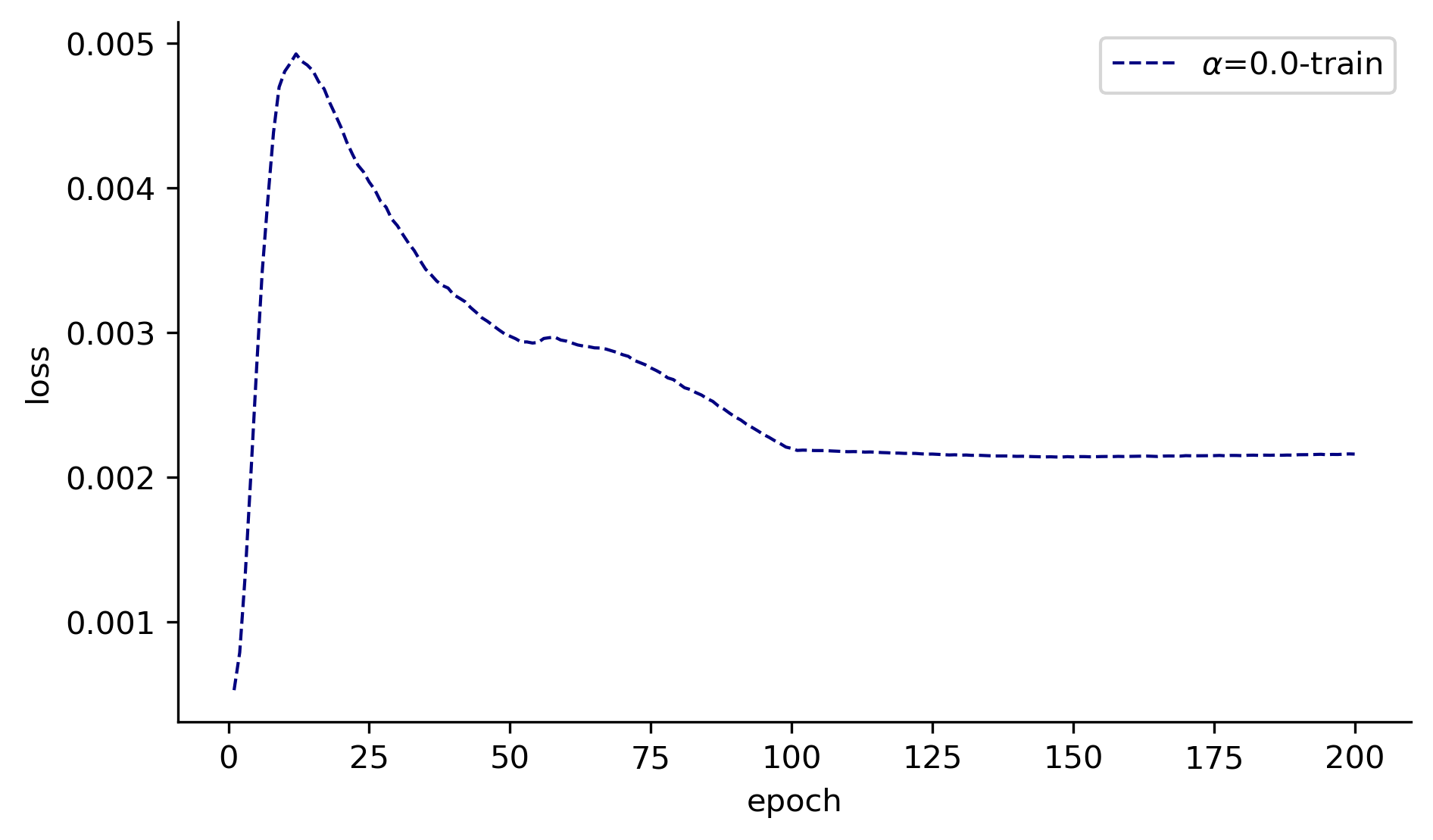}
	\end{subfigure}

	\begin{subfigure}{.22\textwidth}
		\centering
		\includegraphics[width=\textwidth]{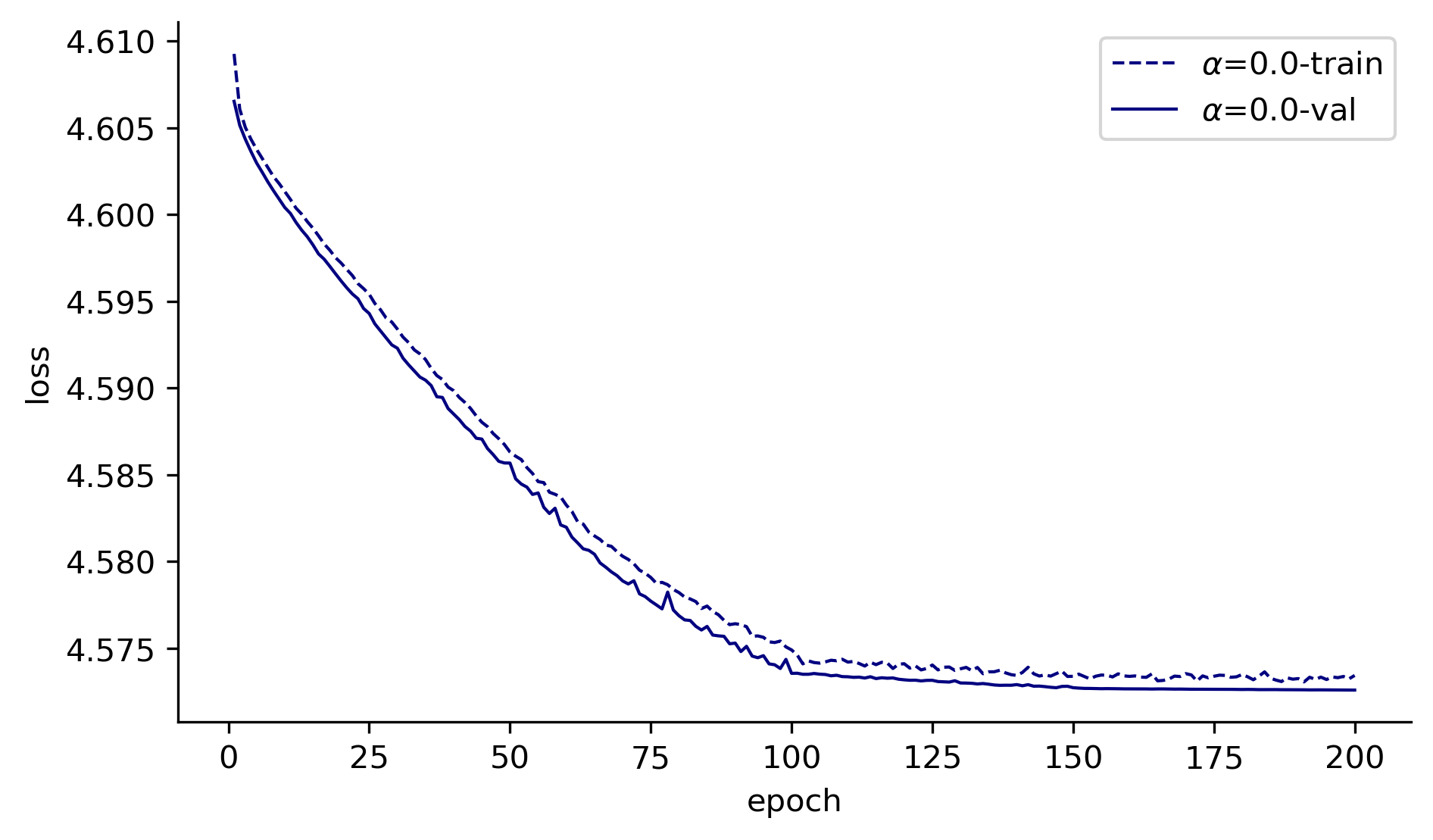}
		\caption{Train and validation loss.}
	\end{subfigure}
	\begin{subfigure}{.22\textwidth}
		\centering
		\includegraphics[width=\textwidth]{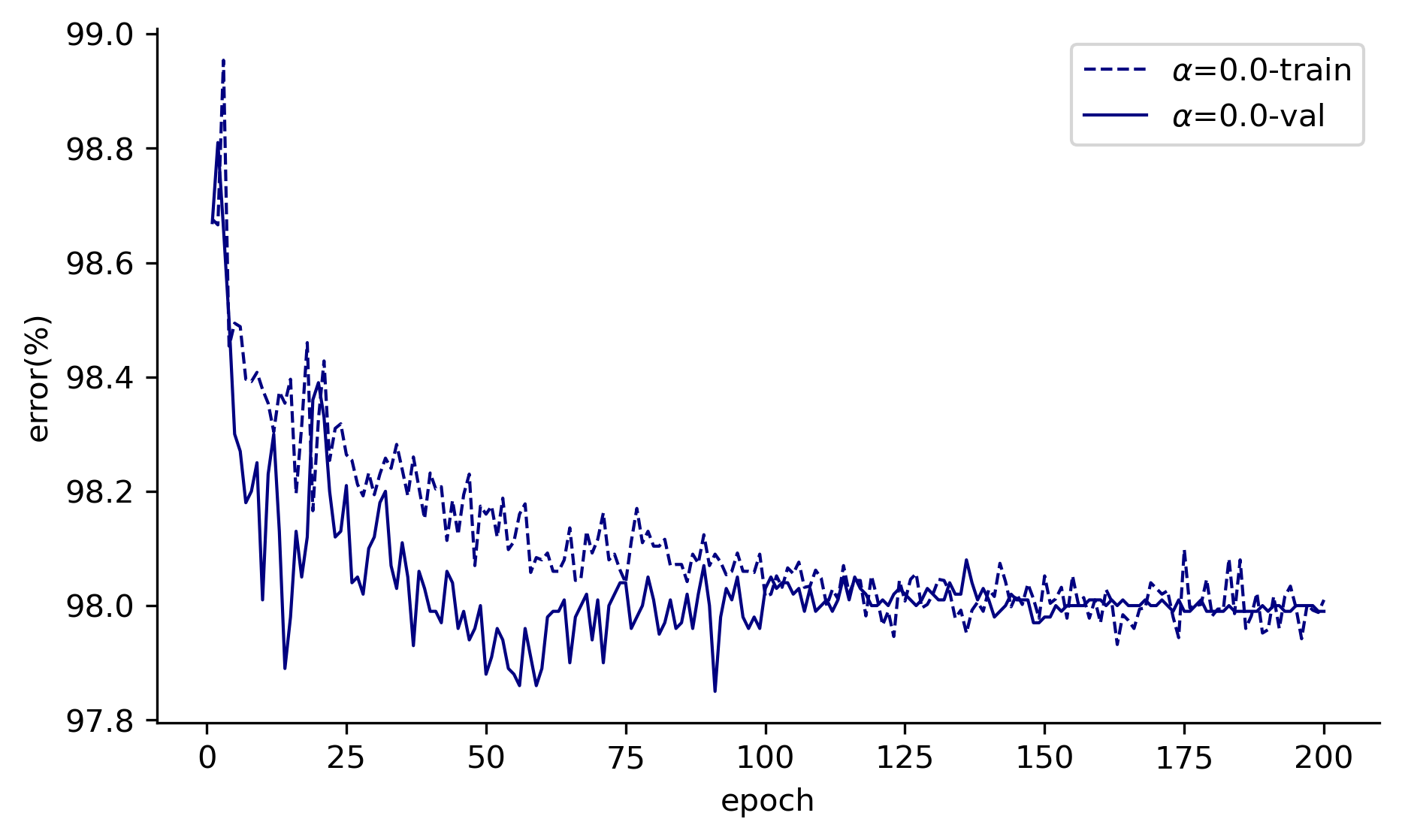}
		\caption{Train and validation error.}
	\end{subfigure}
	\begin{subfigure}{.22\textwidth}
		\centering
		\includegraphics[width=\textwidth]{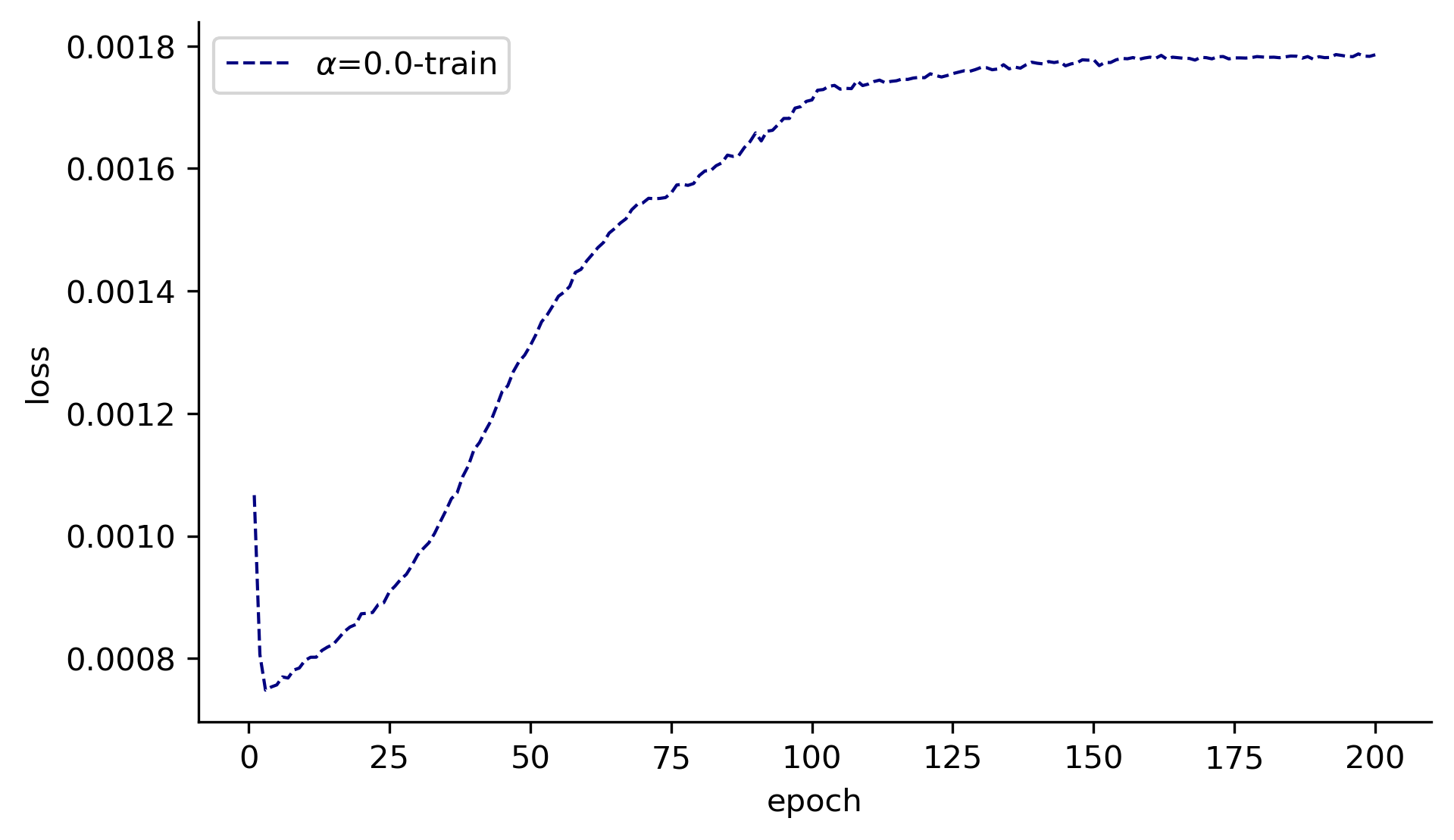}
		\caption{Input loss.}
	\end{subfigure}
	\caption{\textbf{Top:} MLP training on MNIST. \textbf{Middle:} Plain-20 training on CIFAR-10. \textbf{Bottom:} Plain-20 training on CIFAR-100. Setting the $\alpha$ of three passes learning to $0$ seems to converge in the three cases, and it also performs better than the random prediction in the three cases. Moreover, the effects of $\alpha$ of $0$ prove the input loss deserves our attention. The performance on MNIST is the best at the beginning, getting worse as the training goes on and finally converging to some point.}
	\label{mnist alpha0}
\end{figure}


\end{document}